\pdfoutput=1

\documentclass[11pt]{article}

\usepackage[final]{acl}

\usepackage{times}
\usepackage{latexsym}
\usepackage{graphicx}
\usepackage{latexsym}
\usepackage{supertabular}
\usepackage{stfloats}
\usepackage{tabularx}
\usepackage{multirow}
\usepackage{makecell}
\usepackage{booktabs}
\usepackage{amsmath}
\usepackage{amsfonts}
\usepackage{amssymb}
\usepackage{enumerate}
\usepackage{algorithm}
\usepackage{algorithmic}
\usepackage{bbm}
\usepackage{pifont}
\usepackage{colortbl}
\usepackage{xspace}
\usepackage{inconsolata}
\usepackage{subfigure}
\usepackage{caption}
\usepackage{subcaption}
\usepackage{enumitem}

\usepackage{times}
\usepackage{latexsym}
\usepackage{graphicx}  
\usepackage{float}  
\usepackage{subfigure}  

\usepackage[T1]{fontenc}

\usepackage[utf8]{inputenc}

\usepackage{microtype}

\usepackage{inconsolata}

\usepackage{graphicx}

\title{\textit{Less, but Better}: Efficient Multilingual Expansion for LLMs\\via Layer-wise Mixture-of-Experts}

\author{Xue Zhang\textsuperscript{1,2}\thanks{ \ \ This work was done during the internship at Pattern Recognition Center, WeChat AI, Tencent Inc, China.}, Yunlong Liang\textsuperscript{3}, 
Fandong Meng\textsuperscript{3}, Songming           Zhang\textsuperscript{1,2},\\
\textbf{Yufeng Chen\textsuperscript{1,2}\thanks{ \ \ Yufeng Chen is the corresponding author.}},
\textbf{Jinan Xu\textsuperscript{1,2}}, \textbf{Jie Zhou\textsuperscript{3}} \\
\textsuperscript{1}Key Laboratory of Big Data \& Artificial Intelligence in Transportation,\\(Beijing Jiaotong University), Ministry of Education \\
\textsuperscript{2}School of Computer Science and Technology, Beijing Jiaotong University, Beijing, China \\
\textsuperscript{3}Pattern Recognition Center, WeChat AI, Tencent Inc, China \\
\texttt{\{\text{zhang\_xue},smzhang22,chenyf,jaxu\}@bjtu.edu.cn}}

\begin{document}
\maketitle

\begin{abstract}
Continually expanding new languages for existing large language models (LLMs) is a promising yet challenging approach to building powerful multilingual LLMs.
The biggest challenge is to make the model continuously learn new languages while preserving the proficient ability of old languages.
To achieve this, recent work utilizes the Mixture-of-Experts (MoE) architecture to expand new languages by adding new experts and avoid catastrophic forgetting of old languages by routing corresponding tokens to the original model backbone (old experts).
Although intuitive, this kind of method is parameter-costly when expanding new languages and still inevitably impacts the performance of old languages.
To address these limitations, we analyze the language characteristics of different layers in LLMs and propose a layer-wise expert allocation algorithm (\textbf{LayerMoE}) to determine the appropriate number of new experts for each layer.
Specifically, we find different layers in LLMs exhibit different representation similarities between languages and then utilize the similarity as the indicator to allocate experts for each layer, \textit{i.e.}, the higher similarity, the fewer experts.
Additionally, to further mitigate the forgetting of old languages, we add a classifier in front of the router network on the layers with higher similarity to guide the routing of old language tokens.
Experimental results show that our method outperforms the previous state-of-the-art baseline with 60\% fewer experts in the single-expansion setting and with 33.3\% fewer experts in the lifelong-expansion setting, demonstrating the effectiveness of our method.

\end{abstract}

\section{Introduction}

Although existing large language models (LLMs) have exhibited remarkable ability in high-resource languages, their performance in multilingual scenarios is still limited \cite{lai-etal-2023-chatgpt, zhang-etal-2024-enhancing-multilingual, huang2025surveylargelanguagemodels}.
To enhance the multilingual ability of LLMs, continually expanding new languages for existing LLMs is a sustainable approach, which does not require excessive computation or data resources than re-training from scratch \cite{huang2024surveylargelanguagemodels}.
During this process, the key challenge is updating the model with new language data while avoiding catastrophic forgetting of the ability in originally proficient languages.

As a promising solution, MoE-LPR \cite{zhou2024moelprmultilingualextensionlarge}, a two-stage training strategy, utilizes the Mixture-of-Experts (MoE) architecture to achieve expansion of new languages and seeks to preserve the ability of the original languages by targeted routing.
Specifically, at the first stage, it adds and trains several new experts beyond the dense LLM to learn new languages.
Then, at the second stage, it tries to recover the ability of old languages by language priors routing training.
Although this method shows the potential of MoE on balancing the capability of newly expanded languages and the original languages, there are still two problems.
For one thing, the expansion of MoE-LPR is inefficient and leads to high resource usage after expansion (\textit{e.g.}, the original Qwen1.5-1.8B has to be 3.2$\times$ larger to accommodate three additional languages).
For another, although MoE-LPR designs the language priors routing to recover the abilities of the original languages, the corresponding performance of old languages still declines noticeably.

Towards these limitations, we first analyze the similarity between hidden states in different languages and find that some layers exhibit high similarity.
On this basis, we propose a layer-wise expert allocation algorithm (\textbf{LayerMoE}) to determine the appropriate number of new experts for each layer, with the insight that higher representation similarity requires fewer experts.
Additionally, we speculate that high similarity between old and new languages in some layers may confuse the router network, causing the forgetting of old languages. 
To further mitigate the forgetting, we add a classifier in front of the router network on the layers exhibiting higher similarity to guide the routing of tokens in old languages.

We conduct experiments under two settings: (1) single-expansion, where the model is enhanced once by incorporating a new language group, and (2) lifelong-expansion, where the model undergoes sequential adaptation to multiple new language groups.
The corresponding experimental results on four benchmarks demonstrate the effectiveness of our LayerMoE.
Specifically, in the single-expansion setting (please refer to Table  \ref{table:single-res}), our method achieves superior performance with 60\% fewer experts compared to the previous state-of-the-art method, \textit{i.e.}, better expansion of new languages, and less forgetting of old languages.
In the lifelong-expansion setting (please refer to Table \ref{table:lifelong-res}), our method also outperforms baselines, requiring 33.3\% fewer experts.
These results indicate that our expert allocation algorithm offers a resource-efficient solution for continuous language learning, and the classifier added in front of the router network further prevents the forgetting of old languages.

In summary, the major contributions of this paper are as follows\footnote{The code is publicly available at \url{https://github.com/XZhang00/LayerMoE}.}:
\begin{itemize}
    \item We devise a layer-wise expert allocation algorithm to assign the appropriate number of new experts for each layer, which is parameter-efficient and achieves better expansion performance of new languages.

    \item We add a classifier in front of the router network on the layers that exhibit higher similarity, further mitigating the forgetting of old languages.

    \item Experimental results in the single-expansion and lifelong-expansion settings demonstrate the effectiveness and the efficiency of our LayerMoE.
    
\end{itemize}

\section{Background}

In this section, we first introduce the two-stage expansion method MoE-LPR \cite{zhou2024moelprmultilingualextensionlarge} and discuss its limitations.

\paragraph{Stage-1: Continual pre-training with MoE.}
The dense model is upcycled to an MoE model by adding new feed-forward networks (FFNs) and router networks.
Formally, the MoE model includes $N$ experts and a linear router network $W_r \in \mathbb{R}^{h \times N}$ for each layer, where $h$ is the dimension of hidden states after the Attention module.
During the forward process, the top-$K$ experts $\mathcal{I}$ are dynamically selected by the router according to the router scores:
\begin{equation}
    G(\mathbf{x}) = \text{Softmax} (\mathbf{x} \cdot W_r),
\end{equation}
\begin{equation}
    \mathcal{I} = \{i \mid G_i(\mathbf{x}) \in \mathrm{TopK}(G(\mathbf{x}), K)\},
\end{equation}
where $\mathbf{x} \in \mathbb{R}^h$ is the input token hidden state vector of the router network, $G(\mathbf{x}) \in \mathbb{R}^N$ represents the router scores of each expert, and $K$ is the number of the activated experts ($K = 2$).
Then we normalize the weights of the selected experts and weighted sum their outputs to obtain the final representation:
\begin{equation}\label{equation:route}
    \mathbf{y} = \sum_{i \in \mathcal{I}} \frac{G_i(\mathbf{x})}{\Sigma_{j \in \mathcal{I}}G_j(\mathbf{x})} E_i(\mathbf{x}) + \mathbf{x},
\end{equation}
where $G_i(\mathbf{x})$ and $E_i(\mathbf{x})$ denote the router score and the output of the \textit{i}-th expert respectively.

In stage-1, the parameters of the original dense model are frozen to avoid catastrophic forgetting and the newly added experts and routers are updated with the data of expanded languages.
The training objective includes the next token prediction loss $\mathcal{L}_{NTP}$ and the load balance loss $\mathcal{L}_{balance}$ \cite{fedus2022switchtransformersscalingtrillion}.
Given the MoE model $\mathcal{M}$ with the parameters of the newly added experts $\theta_{e}$ and the routers $\theta_{r}$, the $\mathcal{L}_{NTP}$ of one input sequence $\mathbf{t}$ is $\mathcal{L}_{NTP}(\theta_{e}, \theta_{r}) = -\sum_{i}^{|\mathbf{t}|}\log p_{\mathcal{M}}(t_i\mid \mathbf{t}_{<i})$.
The $\mathcal{L}_{balance}$ is an expert-level load balance loss (please refer to Appendix \ref{sec:appendix-load-balande-loss} for the detailed introduction) to mitigate the risk of routing collapse.
The final training loss of stage-1 is:
\begin{equation}
    \mathcal{L}_{stage1}(\theta_{e}, \theta_{r}) = \mathcal{L}_{NTP} + \alpha \mathcal{L}_{balance},
\end{equation}
where $\alpha$ is a hyper-parameter that controls the weight of the load balance loss ($\alpha = 0.01$).

\paragraph{Stage-2: Review with LPR.}
In this stage, only the routers are updated with the mixed data of the old languages and the new languages.
The language priors routing (LPR) loss is designed to route the old language tokens to the original frozen experts (\textit{i.e.}, the expert 0 in each layer).
Formally, given a batch tokens $\mathbb{T}$ including the old language tokens set $\mathbb{T}_{old}$ and the indicator function $F(t)$, the LPR loss $\mathcal{L}_{LPR}$ is defined as:
\begin{equation}
    \mathcal{L}_{LPR}(\theta_{r}) = -\sum_{t \in \mathbb{T}} F(t)\log  G_0(t),
\end{equation}
\begin{equation}\label{equation:ft}
    F(t) = \left\{
    \begin{array}{ll}
        1, & \text{if } t \in \mathbb{T}_{old}, \\
        0, & \text{if } t \notin \mathbb{T}_{old},
    \end{array}
    \right.
\end{equation}
where $G_0(t)$ is the routing score of the expert 0.
The final training loss of stage-2 is:
\begin{equation}
    \mathcal{L}_{stage2}(\theta_{r}) = \mathcal{L}_{NTP} + \beta \mathcal{L}_{LPR},
\end{equation}
where $\beta$ is a hyper-parameter that controls the weight of the LPR loss ($\beta = 0.1$).

\paragraph{Discussions about MoE-LPR.} 
During stage-1, MoE-LPR directly adds uniformly distributed experts for each layer, \textit{i.e.}, each layer adds the same number of new experts.
This expansion paradigm may lead to inefficient expansion and high resource usage, especially in lifelong learning settings. 
Inspired by the findings on language-specific and language-agnostic characteristics of different layers \cite{chen2023journeycenterknowledgeneurons, zhang2024multilingualknowledgeeditinglanguageagnostic}, we raise a question: do all layers need the same number of new experts? 
One intuition is that decreasing the number of experts in some layers may not hurt the overall performance, which motivates us to design an expert allocation algorithm.
During stage-2, although the LPR loss is used to guide the routing of old language tokens, the performance of the old languages still declines noticeably.
This suggests that the routing rules in some layers may be difficult for the router network to learn.



\section{Methodology}

To address the two limitations of MoE-LPR, we first analyze the similarity of hidden states between different languages across different layers.
Then we introduce our layer-wise expert allocation algorithm based on the similarity.
Lastly, we introduce the classifier ahead of the router network in the layers with high similarity to control the routing of the old languages.

\subsection{Similarity Analysis}\label{sec:similarity}

Considering the simplistic structure of the routing network, which essentially performs a linear transformation, the routing results are typically dependent on the input of the router network, i.e., the Hidden States obtained after the Attention module (HSAs). 
Therefore, we analyze the similarity of HSAs in different languages to investigate the potential differences of languages at different layers.

Given two languages $\ell_1$ and $\ell_2$, we randomly sample many examples for each language and perform forward propagation to obtain HSAs of each token at different layers.
Then we randomly select ${Q}$ tokens for each language and take their HSAs to build the candidate set $\mathbf{Q}_{\ell_1}$ and $\mathbf{Q}_{\ell_2}$ for calculating the similarity of these selected tokens in different languages.
For example in layer $i$, the similarity of $\mathbf{Q}_{\ell_1}^i$ and $\mathbf{Q}_{\ell_2}^i$ are obtained by calculating the cosine similarity of any two HSA vectors:
\begin{equation}\label{equation:similarity}
    \mathrm{sim}(\mathbf{Q}_{\ell_1}^i, \mathbf{Q}_{\ell_2}^i) \!= \! \frac{1}{Q^2} \! \sum_{\mathbf{u} \in \mathbf{Q}_{\ell_1}^i} \! \sum_{\mathbf{v} \in \mathbf{Q}_{\ell_2}^i}  \mathrm{cos} (\mathbf{u}, \mathbf{v}),
\end{equation}
where $\mathrm{cos}(\cdot)$ denotes the cosine similarity, and $\mathbf{u}/\mathbf{v} \in \mathbb{R}^h$.
These HSAs of randomly selected tokens contain different contextual information \cite{ethayarajh-2019-contextual}, which simulates the actual input of the router network.

\begin{figure}[t]
    \centering
    \includegraphics[width=\linewidth]{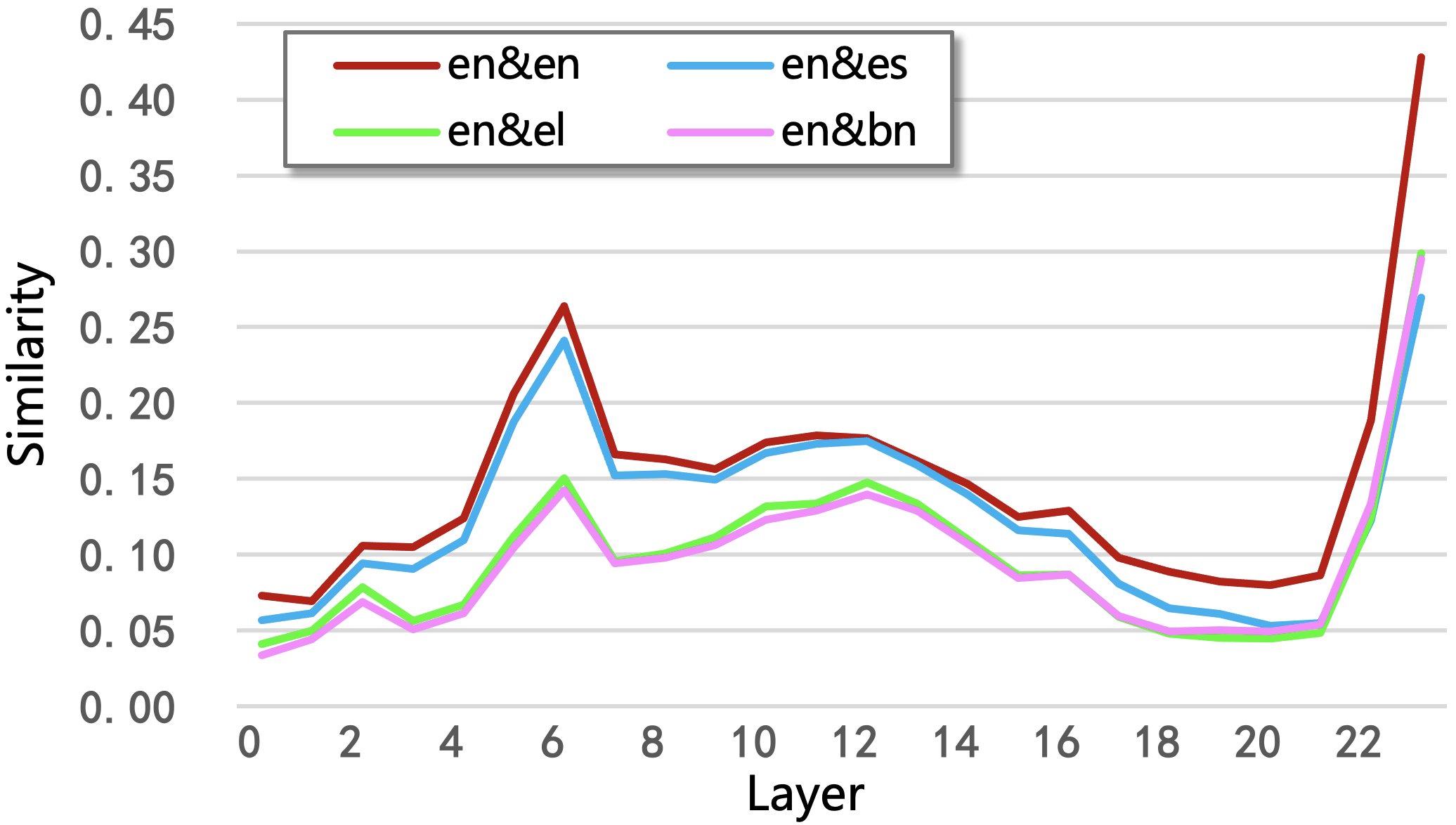}
    \caption{The similarity of HSAs in different languages across all layers within Qwen1.5-1.8B.}
    \label{fig:similarity}
\end{figure}

As a preliminary study, we calculate the similarity of HSAs between some languages in different layers.
We select Qwen1.5-1.8B \cite{qwen1.5} as the base model and choose four languages including English (\textit{en}), Spanish (\textit{es}), Bengali (\textit{bn}), and Greek (\textit{el}).
For each language $\ell$, we sample examples from CulturalX \cite{nguyen-etal-2024-culturax} and then randomly select $Q=100000$ tokens to obtain the candidate set $\mathbf{Q}_{\ell}$.
Then we follow Eq.(\ref{equation:similarity}) to calculate the similarity between four language pairs, including \textit{en}\&\textit{en}, \textit{en}\&\textit{es}, \textit{en}\&\textit{bn}, and \textit{bn}\&\textit{hi}. 
The results in Figure \ref{fig:similarity} show that the similarity in different layers exhibits significant differences, with higher similarity in the middle and last few layers, and lower similarity in the 0\textasciitilde4 layers and 17\textasciitilde21 layers.
\textit{en}\&\textit{en} (the red line) displays the highest similarity across all layers among the four language pairs, while the other 3 language pairs show similar trends with \textit{en}\&\textit{en}.

According to the similarity difference across all layers, we speculate that the higher similarity indicates that HSAs capture more language-agnostic information, while the lower similarity suggests more language-specific characteristics, which aligns with previous studies \cite{zhang2024multilingualknowledgeeditinglanguageagnostic, tang-etal-2024-language} that have observed the existence of language-agnostic/specific neurons within LLMs.
We hypothesize that a layer with higher similarity needs fewer new experts when expanding new languages, as the model can extract language-agnostic representations in this layer. Conversely, lower similarity implies that more experts are necessary to accommodate language-specific features for the new languages.
On this basis, we propose our LayerMoE.

\begin{figure*}[t]
    \centering
    \includegraphics[width=\linewidth]{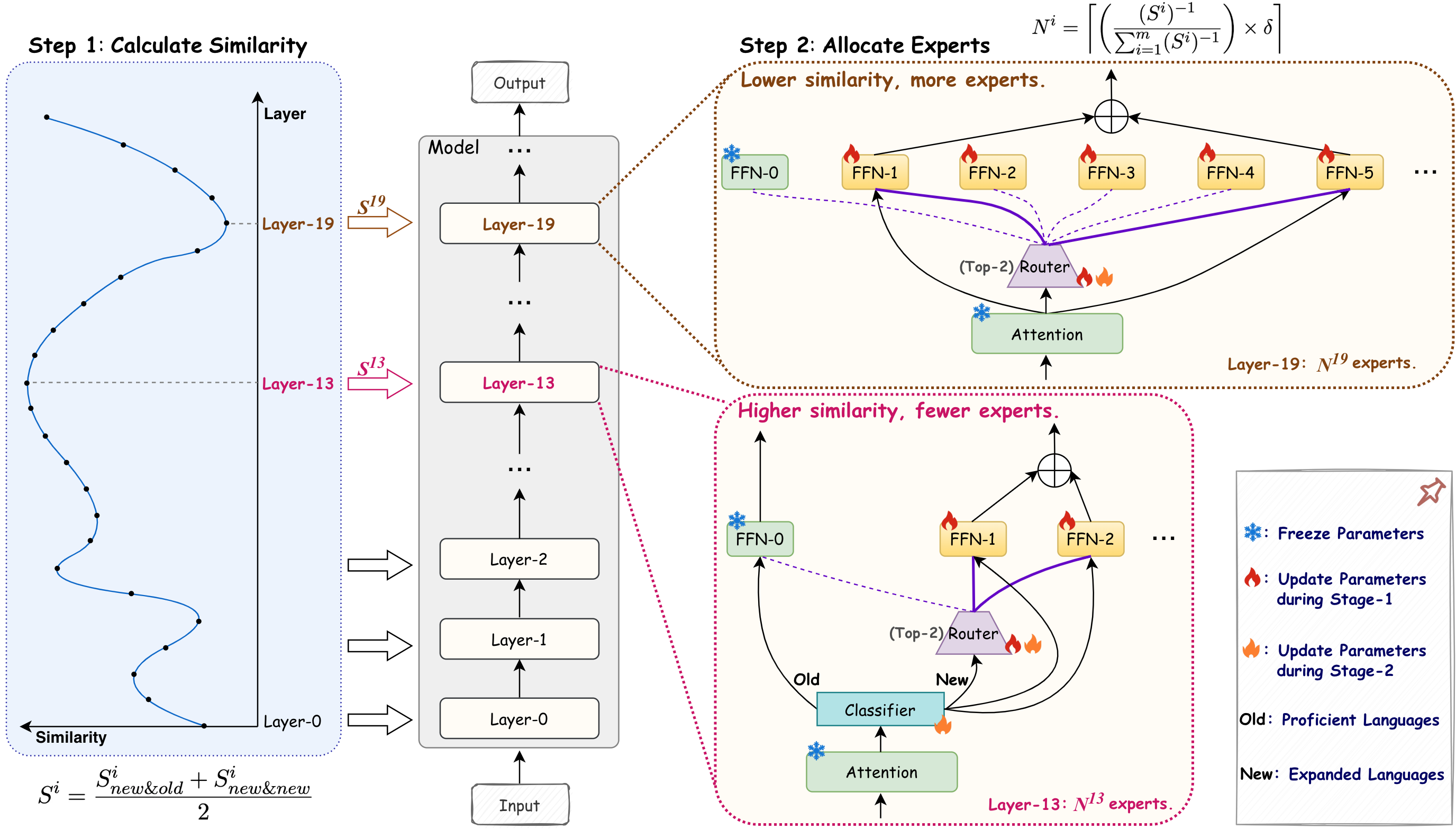}
    \caption{The procedure of our LayerMoE. In Step 1, we first calculate the indicated similarity for each layer. In Step 2, we allocate experts based on the similarity. The layers with lower similarity are allocated more experts, while those with higher similarity are assigned fewer experts. Additionally, we add a classifier in front of the router in the layers with higher similarity to guide the routing of old language tokens.}
    \label{fig:model}
\end{figure*}

\subsection{Layer-wise Expert Allocation}
Figure \ref{fig:model} shows the procedure of our method.
When expanding new languages, we first calculate the similarity between \textit{new}\&\textit{new} and \textit{new}\&\textit{old} languages for each layer.
Then we allocate appropriate experts for each layer according to the similarity.

\paragraph{Step 1: Calculate Similarity.}
Given the dense model $\mathcal{M}_0$, and its proficient language group $L_{old}$ and current expanded language group $L_{new}$, we first randomly select $Q$ tokens to build the candidate token set $\mathbf{Q}_{old}$ and $\mathbf{Q}_{new}$. 
Then we calculate the similarity of $\mathbf{Q}_{new}$\&$\mathbf{Q}_{old}$ and $\mathbf{Q}_{new}$\&$\mathbf{Q}_{new}$, and compute their average as the indicated similarity of each layer.
For example in layer $i$, the indicated similarity $S^i$ is calucluated by:
\begin{equation}\nonumber
    S_{new\&old}^i \! = \!
    \frac{\sum_{j}^{|L_{new}|} \! \sum_{k}^{|L_{old}|} \! \mathrm{sim}(\mathbf{Q}_{\ell_j}^i, \! \mathbf{Q}_{\ell_k}^i)}{|L_{new}||L_{old}|},
\end{equation}
\begin{equation}\nonumber
    S_{new\&new}^i \! = \!
    \frac{\sum_{j}^{|L_{new}|} \!  \sum_{k, k \neq j}^{|L_{new}|}  \! \mathrm{sim}(\mathbf{Q}_{\ell_j}^i,  \! \mathbf{Q}_{\ell_k}^i)}{2|L_{new}|(|L_{new}|-1)},
\end{equation}
\begin{equation}\label{equation:sim-layers}
    S^i = \frac{S_{new\&old}^i + S_{new\&new}^i}{2},
\end{equation}
where $\mathrm{sim}(\cdot, \cdot)$ follows Eq.(\ref{equation:similarity}).

\paragraph{Step 2: Allocate Experts.}
Given the dense model with $m$ layers, we obtain the indicated similarity vector $\mathcal{S} \! = \! [S^1, S^2, \dots, S^m]$ after Step 1.
Then numbers of experts across $m$ layers $\mathcal{N} \! = \! [N^1, N^2, \dots, N^m]$ is calculated as follows:
\begin{equation}\label{equation:expert-number}
    N^i = \left\lceil \left (\frac{({S^i})^{-1}}{\sum_{i=1}^{m}(S^i)^{-1}} \right ) \times \delta \right\rceil,
\end{equation}
where $\delta$ is the target sum number of all experts such that $\sum_{i=1}^m N^i=\delta$, and $\left\lceil \cdot \right\rceil$ denotes round up to the nearest integer.
As a result, fewer experts will be allocated to the layers with higher similarity, while more experts to the layers with lower similarity.
After allocating experts to each layer, we conduct the post-training (stage-1) using the data of new languages to obtain the MoE model $\mathcal{M}_{stage1}$.

\subsection{Routing Classification Network}
During stage-2, the key objective is $\mathcal{L}_{LPR}$ that trains the router network to route the tokens in old languages to old experts.
To reduce the confusion of the routers on old languages and new languages, we add a routing classification network $W_c$ in front of the router network to judge whether the current token belongs to old languages, where $W_c \in \mathbb{R}^{h \times 2}$.
After obtaining the model $\mathcal{M}_{stage1}$, we first calculate the similarities $\mathcal{S}_{new\&old} \! = \![S^1_{new\&old}, S^2_{new\&old}, \dots, S^m_{new\&old}]$ between new and old languages following Eq.(\ref{equation:sim-layers}) for each layer.
Then we add the classifier to the top-$K$ layers according to $\{i | S^i_{new\&old} \in \mathrm{TopK}(\mathcal{S}_{new\&old}, K)\}$.
The training objective of $W_c$ is:
\begin{equation}
    \mathcal{L}_{CLS}(\theta_c) \!=\! -\sum_{t \in \mathbb{T}} F(t)\mathrm{log\_softmax}(\mathbf{x} \cdot W_c),
\end{equation}
where the indicator function $F(t)$ is the same as Eq.(\ref{equation:ft}), $\mathbf{x}$ is the hidden state of the token $t$, and $\theta_c$ denotes the parameters of the classifiers.
Therefore, the final training loss of stage-2 is:
\begin{equation}
    \mathcal{L}_{stage2}(\theta_{r}, \theta_c) = \mathcal{L}_{NTP} + \beta \mathcal{L}_{LPR} + \gamma \mathcal{L}_{CLS},
\end{equation}
where $\beta$ and $\gamma$ are hyper-parameters that control the weights of $\mathcal{L}_{LPR}$ and $\mathcal{L}_{CLS}$.

During inference, the tokens classified to old languages are directly routed to the old expert, while others classified to new languages are normally routed following Eq.(\ref{equation:route}):
\begin{equation}
     \mathbf{y} \!=\! 
\left\{
\begin{array}{ll}
E_0(\mathbf{x}) + \mathbf{x}, \!&\! cls(\mathbf{x})\!=\!0, \\
\sum_{i \in \mathcal{I}} \frac{G_i(\mathbf{x})}{\Sigma_{j \in \mathcal{I}}G_j(\mathbf{x})} E_i(\mathbf{x}) + \mathbf{x}, \!&\! cls(\mathbf{x})\!=\!1,
\end{array}
\right.
\end{equation}
where $cls(\mathbf{x}) = \mathrm{argmax}(\mathbf{x} \cdot W_r)$, ``0/1'' denotes the current token belonging to ``old/new'' languages, and $E_0$ represents the old expert.

\section{Experiments}

\subsection{Experimental Setup}
We conduct the experiments under single-expansion and lifelong-expansion settings. 
The single-expansion setting means expanding one group of new languages to enhance the original model once, while the lifelong-expansion setting means sequentially adapting the model on multiple groups of new languages.

\paragraph{Model and Languages.}
Following MoE-LPR \cite{zhou2024moelprmultilingualextensionlarge}, we select Qwen1.5-1.8B \cite{qwen1.5} as our base model, which has a powerful multilingual tokenizer and lower computation overhead for easily upcycling.
Based on the ability of Qwen1.5-1.8B, we select three languages with high performance as the old language group (G0) to observe the catastrophic forgetting phenomenon: English (\textit{en}), Chinese (\textit{zh}), and Spanish (\textit{es}).
Additionally, we choose six low-resource languages\footnote{The performance on six languages is shown in Table \ref{table:single-res}. More details of each language are listed in Appendix \ref{sec:appendix-languages}.} in which Qwen1.5-1.8B performs poorly as the expanded languages: Greek (\textit{el}), Hungarian (\textit{hu}), Turkish (\textit{tr}), Bengali (\textit{bn}), Hindi (\textit{hi}), and Nepali (\textit{ne}).
Specifically, the selection of \textit{el}, \textit{hu}, and \textit{tr} follows MoE-LPR \cite{zhou2024moelprmultilingualextensionlarge}, while \textit{bn}, \textit{hi}, and \textit{ne} are selected to observe the performance of different methods on morphologically rich non-Latin scripts, which pose greater modeling challenges and suffer from limited resources.
Hence, we divide them into two groups (G1: \textit{el}, \textit{hu}, and \textit{tr}, G2: \textit{bn}, \textit{hi}, and \textit{ne}) for the single-expansion setting (G0 \!$\rightarrow$\! G1, G0 \!$\rightarrow$\! G2, and G0 \!$\rightarrow$\! G1 \!$+$\! G2) and the lifelong-expansion setting (G0 \!$\rightarrow$\! G1 \!$\rightarrow$\! G2, and G0 \!$\rightarrow$\! G2 \!$\rightarrow$\! G1).

\paragraph{Training.} 
During stage-1, we train the new experts and the router network with new language data.
For each new language, we sample 2 billion tokens from CulturalX \cite{nguyen-etal-2024-culturax}, a substantial multilingual dataset with 6.3 trillion tokens in 167 languages.
The detailed training setup follows MoE-LPR \cite{zhou2024moelprmultilingualextensionlarge}: the batch size is 512, the sequence length is 1024, the learning rate is 5e-5, and the weight $\alpha$ of the load balancing loss is 0.01.
During stage-2, we mix the data of old languages and new languages to train the router network and the classification network.
The top-7 layers with high similarity for the single-expansion settings and top-5 layers for the lifelong-expansion settings are taken according to the best performance to add the routing classification network.
We randomly sample 50K examples for each old language and 100K examples for each new language as the training data of stage-2.
Following MoE-LPR \cite{zhou2024moelprmultilingualextensionlarge}, the English/Chinese/Spanish data is sampled from Slimpajam \cite{cerebras2023slimpajama}, SkyPile-150B \cite{wei2023skyworkopenbilingualfoundation}, and CulturalX \cite{nguyen-etal-2024-culturax}, respectively.
The 100k examples for each new language are sampled from the training data of stage-1.
We set the batch size to 768 and the learning rate to 5e-5.
The weight of $\mathcal{L}_{LPR}$ and $\mathcal{L}_{CLS}$ loss is set to 0.1.
All experiments are conducted on 8$\times$A100-40G GPUs.

\paragraph{Evaluation.} 
We evaluate all methods on 4 benchmarks: ARC-Challenge (25-shot) \cite{clark2018thinksolvedquestionanswering}, MMLU (5-shot) \cite{hendrycks2021measuringmassivemultitasklanguage}, HellaSwag (10-shot) \cite{zellers-etal-2019-hellaswag}, and Belebele (5-shot) \cite{Bandarkar_2024}.
We utilize the \texttt{lm-evaluation-hardness}\footnote{\url{https://github.com/EleutherAI/lm-evaluation-harness}} framework \cite{eval-harness} to conduct evaluation on their multilingual version.
Please refer to Appendix \ref{sec:appendix-eval} for the detailed introduction.

\subsection{Baselines}
{MoE-LPR} \cite{zhou2024moelprmultilingualextensionlarge} adds the same number of new experts for each layer. 
\citeauthor{gao2024higherlayersneedlora} proposes MOLA that experts in lower layers are more similar than those in higher layers, \textit{i.e.}, higher layers need more experts.
For G0 \!$\rightarrow$\! G1, G0 \!$\rightarrow$\! G2, we conduct three versions of MoE-LPR, adding 2/3/5 experts per layer as 3*24/4*24/6*24 (Qwen1.5-1.8B contains 24 layers).
And MOLA ($\nabla$, \textbf{2468}) is employed following the best setting, allocating 2/4/6/8 experts for 0\textasciitilde5/6\textasciitilde11/12\textasciitilde17/18\textasciitilde23 layers.
For G0 \!$\rightarrow$\! G1 \!$+$\! G2, G0 \!$\rightarrow$\! G1 \!$\rightarrow$\! G2, and G0 \!$\rightarrow$\! G2 \!$\rightarrow$\! G1, we conduct MoE-LPR (7*24/4*24) as baselines.
For a fair comparison, we reproduce all versions of MoE-LPR and MOLA using the same data as our method, \textit{i.e.}, 2 billion tokens for each new language. 

Additionally, we directly take five baselines of MoE-LPR (trained with 8 billion tokens for each new language) under G0 \!$\rightarrow$\! G1 for a clear comparison, including {LoRA} \cite{hu2021loralowrankadaptationlarge}, {Full Fine-tuning}, {LLaMA-Pro} \cite{wu2024llamaproprogressivellama}, {MoE}, and {LoRAMoE} \cite{dou2024loraMoEalleviateworldknowledge}. 
\textbf{LoRA} \cite{hu2021loralowrankadaptationlarge} introduces Low-Rank Adaptation for efficiently fine-tuning large language models. In this paper, the LoRA targets include all linear modules. The LoRA rank is set to 8.
\textbf{Full Fine-tuning} is fine-tuning all parameters on the dense model.
\textbf{LLaMA-Pro} \cite{wu2024llamaproprogressivellama} insert new transformer blocks to replace the duplicate layers. Only the newly inserted blocks are trained to store new knowledge during post-pretraining. 12 new layers (the best setting) are inserted into the original dense model.
\textbf{MoE} means that only the stage-1 of MoE-LPR is conducted.
\textbf{LoRAMoE} \cite{dou2024loraMoEalleviateworldknowledge} is a novel framework that combines multiple LoRAs with a router network. The router selects all LoRAs for each token. 
The LoRA rank is set to 180, and the number of LoRAs is 8 for each module.
Due to the poor performance of the 5 baselines, we do not reproduce them and directly take the results trained with 8 billion tokens for each new language from the MoE-LPR paper \cite{zhou2024moelprmultilingualextensionlarge}.

\subsection{Allocation of Experts}
\begin{figure}[t]
    \centering
    \includegraphics[width=\linewidth]{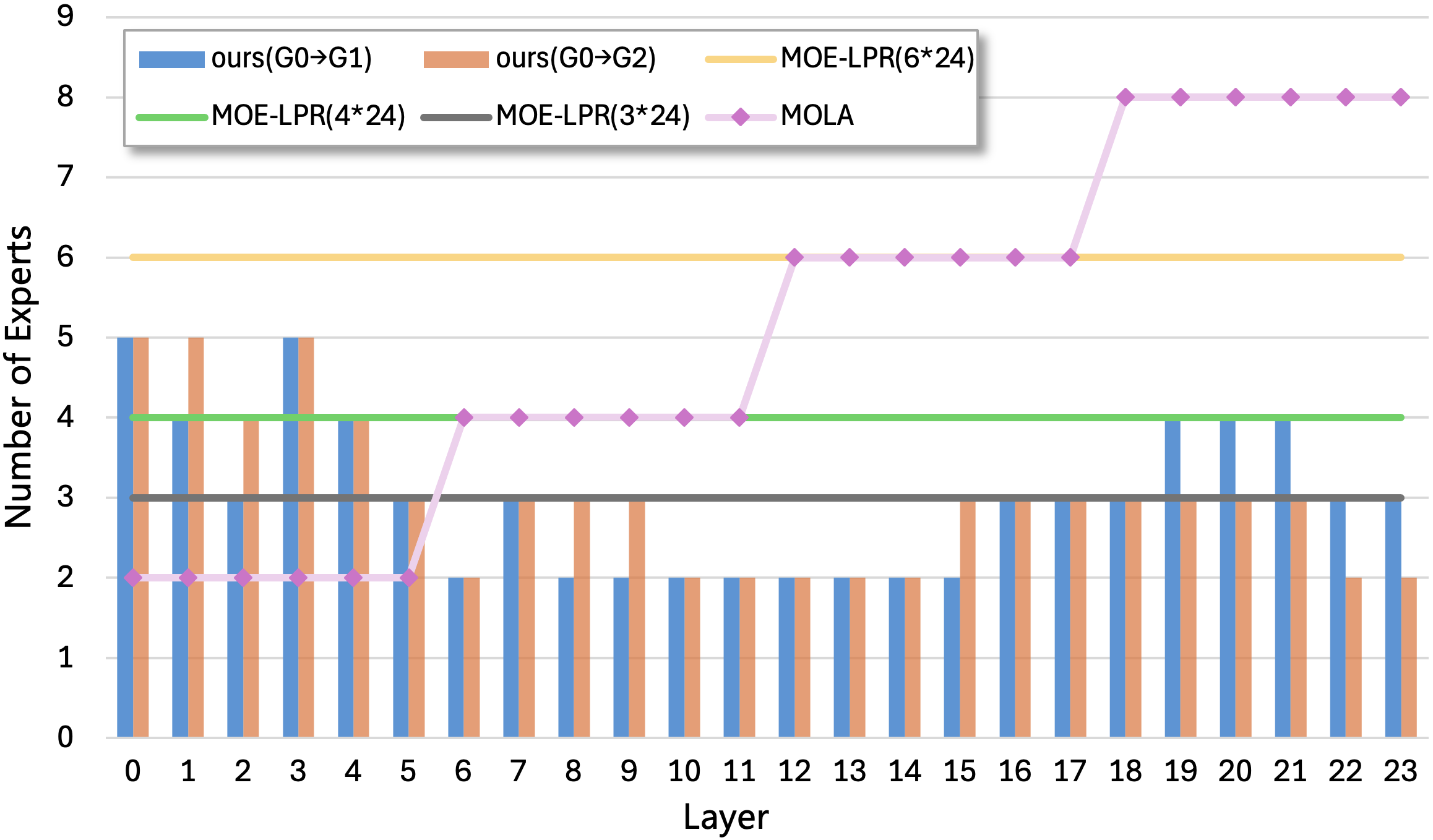}
    \caption{The expert allocation of our method and baselines for each layer under G0 \!$\rightarrow$\! G1, G0 \!$\rightarrow$\! G2.}
    \label{fig:allocation}
\end{figure}

After calculating the similarity following Step 1, we assign the number of new experts for each layer according to Eq.(\ref{equation:expert-number}).
For a fair comparison with MoE-LPR (3*24), we set the targeted number of new experts $\delta$ in Eq.(\ref{equation:expert-number}) to 72.
The allocation results of our LayerMoE and baselines (under G0 \!$\rightarrow$\! G1 and G0 \!$\rightarrow$\! G2) are present in Figure \ref{fig:allocation}.
Our method allocates more experts for the shallow and deep layers and less experts for the middle layers.
The detailed similarity and expert allocation results of our method under each setting are plotted in Figure \ref{fig:sim_experts} of Appendix \ref{sec:appendix-sim_and_allocation}.

\begin{table*}[t]
    \centering
    \resizebox*{\linewidth}{!}{
    \begin{tabular}{lc|cccccccc|ccc}
    \bottomrule
        & & \multicolumn{2}{c}{\textbf{ARC-C}} & \multicolumn{2}{c}{\textbf{MMLU}} & \multicolumn{2}{c}{\textbf{Hellaswag}} & \multicolumn{2}{c|}{\textbf{Belebele}} & & & \\
        \hline
        \textbf{Methods} & \textbf{Expand-params} &  \textbf{Old} &	\textbf{New} &	\textbf{Old}  &	\textbf{New} & \textbf{Old} &	\textbf{New}	& \textbf{Old}	&  \textbf{New} &	\textbf{\textit{Old-avg}} & \textbf{\textit{New-avg}} & \textbf{\textit{avg}}\\
        
    \toprule
    \multicolumn{13}{c}{\textbf{Old (G0): \textit{en}, \textit{es}, \textit{zh} \!$\rightarrow$\! New (G1): \textit{el}, \textit{hu}, \textit{tr}}} \\
    \bottomrule
    Qwen1.5-1.8B	&	/	&	33.31 	&	22.93 	&	47.23 	&	30.52 	&	49.70 	&	29.27 	&	55.89 	&	32.92 	&	46.53 	&	28.91 	&	37.72 	\\
    \hline
    $^*$LoRA	&	/ & 28.33 &	23.89 &	37.42 & 29.30 & 41.48 & 29.78 & 39.45 & 26.93 & 36.67	& 27.48	& 32.07 \\
    $^*$Full Fine-tuning	& / &	31.72 & 25.98 & 43.51 & 33.18 & 47.38 & 35.28 & 45.26 & 33.70 & 41.98 &	32.04 & 	37.00 \\
    $^*$LLaMA-Pro	& 0.6B &	31.77 & 24.35 & 44.06 & 34.02 & 48.36 & 33.85 & 48.78 & 31.52 & 43.24 &	30.94 & 	37.09 \\
    $^*$MoE	 & 4B	& 32.51 & 26.43 & 44.16 & 35.07 & 48.54 & 37.01 & 45.37 & 32.74 & 42.65 &	32.81 &	37.73 \\
    $^*$LoRAMoE	 & 0.8B &	32.43 & 26.63 & 45.41 & 34.17 & 48.61 & 37.17 & 47.74 & 32.81 & 43.55	& 32.70 &	38.12 \\
    \hline
    \rowcolor{gray!25}
    MoE-LPR (6*24)	&	4B	&	32.26 	&	26.55 	&	46.51 	&	34.95 	&	49.75 	&	37.53 	&	52.92 	&	39.04 	&	45.36 	&	34.52 	&	39.94 	\\
    
    MOLA ($\nabla$)	&	3.2B ($\downarrow$ 20\%)	&	32.49 	&	27.00 	&	46.38 	&	33.98 	&	49.63 	&	37.03 	&	53.85 	&	38.37 	&	45.59 	&	34.10 	&	39.84 	\\
    
    MoE-LPR (4*24)	&	2.4B ($\downarrow$ 40\%)	&	32.29 	&	26.69 	&	46.51 	&	34.11 	&	49.60 	&	37.12 	&	52.33 	&	37.74 	&	45.18 	&	33.92 	&	39.55 	\\
    
    MoE-LPR (3*24)	&	1.6B ($\downarrow$ 60\%)	&	32.77 	&	26.43 	&	46.54	&	34.59 	&	49.63 	&	37.08 	&	52.07 	&	37.15 	&	45.25 	&	33.81 	&	39.53 	\\
    
    LayerMoE (Ours)	&	\textbf{1.6B ($\downarrow$ 60\%)}	&	\textbf{33.03} 	&	\textbf{26.66} 	&	46.12 	&	34.80 	&	49.24 	&	37.01 	&	\textbf{54.82} 	&	\textbf{40.37}	&	\textbf{45.80} 	&	\textbf{34.71} 	&	\textbf{40.26} 	\\
    \toprule
    
    \multicolumn{13}{c}{\textbf{Old (G0): \textit{en}, \textit{es}, \textit{zh} \!$\rightarrow$\! New (G2): \textit{bn}, \textit{hi}, \textit{ne}}} \\
    \bottomrule
    Qwen1.5-1.8B	&	/	&	33.31 	&	23.05 	&	47.23 	&	29.54 	&	49.70 	&	28.22 	&	55.89 	&	28.85 	&	46.53 	&	27.41 	&	36.97 	\\
    \hline
    \rowcolor{gray!25}
    MoE-LPR (6*24)	&	4B	&	33.06 	&	23.85 	&	47.04 	&	30.62 	&	49.69 	&	30.92 	&	53.11 	&	33.85 	&	45.72 	&	29.81 	&	37.77 	\\
    
    MOLA ($\nabla$)	&	3.2B ($\downarrow$ 20\%)	&	32.48 	&	23.67 	&	47.02 	&	30.82 	&	49.73 	&	30.48 	&	53.59 	&	32.15 	&	45.71 	&	29.28 	&	37.49 	\\

    MoE-LPR (4*24)	&	2.4B ($\downarrow$ 40\%)	&	33.08 	&	23.27 	&	47.21 	&	30.74 	&	49.64 	&	30.67 	&	52.85 	&	33.04 	&	45.70 	&	29.43 	&	37.56 	\\
    
    MoE-LPR (3*24)	&	1.6B ($\downarrow$ 60\%)	&	32.77 	&	23.53 	&	46.97 	&	30.72 	&	49.68 	&	30.56 	&	52.59 	&	33.33 	&	45.50 	&	29.53 	&	37.52 	\\

    LayerMoE (Ours)	&	\textbf{1.6B ($\downarrow$ 60\%)}	& 32.52 &	\textbf{24.02} &	\textbf{47.10} & 	\textbf{30.85} & 	49.46 & 	30.47 &	\textbf{55.15} &	\textbf{34.11} & \textbf{46.06} & \textbf{29.86} & \textbf{37.96} \\

    \toprule
    \end{tabular}
    }
    \caption{
        The results of the single-expansion settings (G0 \!$\rightarrow$\! G1, G0 \!$\rightarrow$\! G2). The results marked with $^*$ represent that they are directly taken from the MoE-LPR paper \cite{zhou2024moelprmultilingualextensionlarge}, which utilizes 8 billion tokens for each new language during post-pretraining.
        Other experiments are trained with 2 billion tokens for each new language.
        ``Expand-params'' denotes the parameter number of newly added experts and {($\downarrow$ *\%)} means the reduction percentage of parameters compared to MoE-LPR (6*24).
        The results in \textbf{bold} mean they are better than the previous SOTA method ``MoE-LPR (6*24)''.
        The detailed results for each language are reported in Table \ref{table:single-res-detailed-g1} and \ref{table:single-res-detailed-g2} of Appendix \ref{sec:appendix-detailed-results}.
    }
    \label{table:single-res}
\end{table*}

\section{Results and Analysis}

\subsection{Main Results}

\paragraph{Single-Expansion.}
The evaluation results of the single-expansion setting  (G0 \!$\rightarrow$\! G1 and G0 \!$\rightarrow$\! G2) are listed in Table \ref{table:single-res}.
The results show that our method surpasses all baselines in ``{\textit{Old-avg}}'' and ``{\textit{New-avg}}'' under the expansion of G1 and G2.
Particularly, our method outperforms the previous SOTA method ``MoE-LPR (6*24)'' with 60\% fewer parameter numbers of newly added experts (\textit{i.e.}, from 4B to 1.6B), which suggests that our method is more parameter-efficient.
On the contrary, MOLA and MoE-LPR (3*24/4*24) underperform MoE-LPR (6*24) when decreasing the number of expansion parameters.
Although other baselines (LoRA, Full Fine-tuning, LLaMA-Pro, MoE, and LoRAMoE) utilize 8B tokens for each new language during post-pretraining, they still perform poorly due to the lack of special designs.
Additionally, when keeping the same parameter numbers (1.6B) with ``MoE-LPR (3*24)'', our method exhibits a significant improvement over ``MoE-LPR (3*24)'' referring to  ``\textit{New-avg}'' (for G1 from 33.81 to 34.71 and for G2 from 29.53 to 29.86).
This superiority indicates that allocating experts based on the language similarity in different layers has a positive contribution to the expansion of new languages.
Furthermore, the results of ``{\textit{Old-avg}}' showcase that our method performs the best on preserving the ability of the old languages, with 45.80 for expanding G1 and 46.06 for G2.
These results prove that appropriate correction of the router network can effectively mitigate the catastrophic forgetting of old languages.

The expansion performance exhibits differences among different languages according to ``\textit{New-avg}''.
When conducting G0 \!$\rightarrow$\! G1, the performance of new languages improves by 5.8 from 28.91 to 34.71, while only improves by 2.45 (from 27.41 to 29.86) on G0 \!$\rightarrow$\! G2.
Although the amounts of training tokens (2B) for each new language are the same, the morphologically rich non-Latin languages in G2 (\textit{bn}, \textit{hi}, \textit{ne}) obtain a smaller performance improvement compared to expanding G1. 
The difference can be attributed to their lower encoding efficiencies \cite{arnett2024languagemodelsperformworse, rust-etal-2021-good, ahia-etal-2023-languages, petrov2023languagemodeltokenizersintroduce, ali-etal-2024-tokenizer}, \textit{i.e.}, these languages typically employ UTF-8 encoding during the tokenization process, which results in longer encoded sequences. 
Consequently, the number of effective training tokens is reduced, limiting the potential for performance improvement.
These results suggest that more tokens are needed for better expansion of G2.
Furthermore, when expanding G1 or G2, there are different degrees of forgetting of old languages.
Expanding G1 brings more forgetting (from 46.53 to 45.80) compared to expanding G2 (from 46.53 to 46.06), which may be caused by more token overlaps between G0 and G1 since the writing systems of \textit{en/es} in G0 and \textit{hu/tr} in G1 all contain the Latin alphabet (as shown in Table \ref{table:languages} of Appendix \ref{sec:appendix-languages}).

\begin{table*}[t]
    \centering
    \resizebox*{\linewidth}{!}{
    \begin{tabular}{lcccc|cccc}
    \bottomrule
        \textbf{Methods} & \textbf{Order} & \textbf{G1-E-p} &  \textbf{G2-E-p} & \textbf{All-E-p} & \textbf{\textit{G0-avg}} & \textbf{\textit{G1-avg}} & \textbf{\textit{G2-avg}} & \textbf{\textit{avg}} \\
    \hline 
    Qwen1.5-1.8B	& / & / & / & / & 46.53	& 28.91 & 27.41 & 34.29	\\
    \hline
    MoE-LPR (7*24) & G0 \!$\rightarrow$\! G1 \!$+$\! G2 & 4.8B & 4.8B & 4.8B   &	45.15 &	34.17 &	29.98 & 36.43 \\
    MoE-LPR (4*24) & G0 \!$\rightarrow$\! G1 \!$+$\! G2 & 2.4B & 2.4B & 2.4B ($\downarrow$ 50.0\%)  & 45.28 &	34.07 &	29.55 & 36.30 \\
    LayerMoE (Ours) & G0 \!$\rightarrow$\! G1 \!$+$\! G2 & 1.6B & 1.6B & \textbf{1.6B ($\downarrow$ \textbf{66.7\%})}  & \textbf{45.56} &	\textbf{34.73} &	29.94 & \textbf{36.74} \\
    
    \hline
    MoE-LPR (7*24) & G0 \!$\rightarrow$\! G1 \!$\rightarrow$\! G2 & 2.4B & 2.4B & 4.8B & 45.88	& 34.40	& 30.26 & 36.85 \\
    LayerMoE (Ours) & G0 \!$\rightarrow$\! G1 \!$\rightarrow$\! G2 & 1.6B  & 1.6B & \textbf{3.2B ($\downarrow$ \textbf{33.3\%})} &  \textbf{46.10} &	\textbf{34.56} &	\textbf{30.29} & \textbf{36.98}  \\
    \hline
    MoE-LPR (7*24) & G0 \!$\rightarrow$\! G2 \!$\rightarrow$\! G1 & 2.4B & 2.4B & 4.8B & 45.28  &	33.42 &	28.39 & 35.70 \\
    LayerMoE (Ours) & G0 \!$\rightarrow$\! G2 \!$\rightarrow$\! G1 & 1.6B  & 1.6B  & \textbf{3.2B ($\downarrow$ \textbf{33.3\%})} & \textbf{46.25}  & \textbf{33.76} &	\textbf{29.80} & \textbf{36.60} \\

    \toprule
    \end{tabular}
    }
    \caption{
        The results of G0 \!$\rightarrow$\! G1 \!$+$\! G2 and the lifelong-expansion settings (G0 \!$\rightarrow$ \!G1 \!$\rightarrow$ \!G2 and G0 \!$\rightarrow$ \!G2 $\rightarrow$ \!G1).
        ``G1/G2/All-E-p'' represents the expansion number of newly added parameters for G1, G2, and the sum, respectively.
        The detailed results for each language are reported in Table \ref{table:mixed-res-detailed} of Appendix \ref{sec:appendix-detailed-results}.
    }
    \label{table:lifelong-res}
\end{table*}

\paragraph{Lifelong-Expansion.}
The results of G0 \!$\rightarrow$ \!G1 \!$\rightarrow$ \!G2 and G0 \!$\rightarrow$ \!G2 $\rightarrow$ \!G1 in Table \ref{table:lifelong-res} demonstrate that our method outperforms MoE-LPR (7*24) with fewer 33.3\% new experts in terms of ``\textit{G0/G1/G2-avg}'', proving the effectiveness of our method in lifelong-expansion setting.
Additionally, we surprisingly observe that G0 \!$\rightarrow$ \!G1 \!$\rightarrow$ \!G2 brings better expansion performance for G2 (``\textit{G2-avg}'') and less forgetting of old languages (``\textit{G0-avg}'') compared to G0 \!$\rightarrow$\! G1 \!$+$\! G2.
But G0 \!$\rightarrow$ \!G2 \!$\rightarrow$ \!G1 does not show similar improvement for G1.
This difference indicates that the learning order of different languages may have non-trivial influences on the final performance of each language.

\begin{table}[t]
    \centering
    \resizebox*{\linewidth}{!}{
    \begin{tabular}{l|ll|c}
    \bottomrule
        	& \textbf{\textit{Old-avg} ($\uparrow$)} & \textbf{\textit{New-avg} ($\uparrow$)} & \textbf{\textit{avg} ($\uparrow$)}\\
        
    \hline

    LayerMoE (G0 \!$\rightarrow$\! G1)	&	\textbf{45.80} 	&	\textbf{34.71} 	&	\textbf{40.26} 	\\
    \quad \textit{w/} random & 45.68 ($\downarrow$ 0.12) & 34.11 ($\downarrow$ \textbf{0.60}) & 39.90 \\
    \quad \textit{w/o} classifier & 45.33 ($\downarrow$ \textbf{0.47}) & 34.61 ($\downarrow$ 0.10) & 39.97 \\
\hline
    LayerMoE (G0 \!$\rightarrow$\! G2)	&	\textbf{46.06} 	&	\textbf{29.86} 	&	\textbf{37.96} 	\\
    \quad \textit{w/} random & 45.94 ($\downarrow$ 0.12) & 29.15 ($\downarrow$ \textbf{0.71}) & 37.54 \\
    \quad \textit{w/o} classifier & 45.39 ($\downarrow$ \textbf{0.67}) & 29.53 ($\downarrow$ 0.33) & 37.46 \\
    
    \toprule
    \end{tabular}
    }
    \caption{
        The ablation results under G0 \!$\rightarrow$\! G1 and G0 \!$\rightarrow$\! G2. ``\textit{w/} random'' means randomly allocating numbers of new experts for each layer without calculating the similarity. ``\textit{w/o} classifier'' denotes no classifier is added in front of the router network.
    }
    \label{table:ablation-study}
\end{table}

\subsection{Ablation Study}

To verify the effectiveness of our expert allocation algorithm and the added classifier, we conduct the ablation experiments under G0 \!$\rightarrow$\! G1 and G0 \!$\rightarrow$\! G2 settings and report the results in Table \ref{table:ablation-study}.
These results show that randomly allocating experts for each layer without calculating the similarity significantly decreases the expansion performance ($\downarrow$ 0.60 in G1, $\downarrow$ 0.71 in G2).
Additionally, the forgetting of old languages becomes worse ($\downarrow$ 0.47 in G1, $\downarrow$ 0.67 in G2) when no classifiers are added in front of the router network.
Overall, these results prove the necessity of our layer-wise expert allocation algorithm and the added classifier.

\subsection{Classifiers in Different Layers}
The top-$K$ layers with the higher similarity between new and old languages are selected according to $\{i | S^i_{new\&old} \in \mathrm{TopK}(\mathcal{S}_{new\&old}, K)\}$ to add the classifier.
To investigate the best value of $K$, we conduct the experiments from top-1 to top-11 and top-24 (all layers) under G0 \!$\rightarrow$\! G1 and list the results in Table \ref{table:layers-calssifier}.
The results show that as more classifiers are added, the performance of the new language expansion first increases and then decreases, and the performance of the old language preservation increases constantly.
The best overall performance peaks on the top-7 setting.
Additionally, we also report the results of ``random-7'' and ``last-7'' to verify the effectiveness of adding classifiers in layers with higher similarity.
Randomly selecting layers or opting for those with lower similarity consistently yields poor performance in preserving the capability of old languages.

\begin{table}[t]
    \centering
    \resizebox*{\linewidth}{!}{
    \begin{tabular}{l|ccc|l|ccc}
    \bottomrule
        	& \textbf{\textit{Old} } & \textbf{\textit{New} } & \textbf{\textit{avg}} & & \textbf{\textit{Old} } & \textbf{\textit{New} } & \textbf{\textit{avg}} \\
        
    \hline
    top-1 & 44.90 & 34.30 & 39.60 & \textbf{top-7} & 45.80 & 34.71 & \textbf{40.26} \\
    top-2 & 44.81 & 34.69 & 39.75 & top-8 & 45.79 & 34.54 & 40.17 \\
    top-3 & 45.10 & 34.41 & 39.76 & top-9 & 45.95 & 34.42 & 40.19 \\
    top-4 & 45.41 & 34.67 & 40.04 & top-10 & \textbf{46.16} & 34.28 & 40.22 \\
    top-5 & 45.28 & 34.66 & 39.97 & top-24 & 46.04 & 34.32 & 40.18 \\
    top-6 & 45.74 & \textbf{34.72} & 40.23 & top-23 & 46.15 & 33.23 & 39.69 \\
    \hline
    last-7 & 44.96 & 34.60 & 39.78 & random-7 & 45.32 & 34.60 & 39.96 \\
    
    \toprule
    \end{tabular}
    }
    \caption{
        The results of adding classifiers in different layers under G0 \!$\rightarrow$\! G1.
        ``top-*'' denotes adding classifiers to * layers with the highest similarity.
        ``random-7'' means randomly selecting 7 layers, while ``last-7'' represents the 7 layers with the lowest similarity.
    }
    \label{table:layers-calssifier}
\end{table}

\subsection{Generalization Study}
To evaluate the generalization of our method on different models, we conduct a brief verification of our method on G0 \!$\rightarrow$\! G1 with Llama-3.2-3B \cite{grattafiori2024llama3herdmodels} and evaluate the performance on 5 benchmarks. 
Particularly, we add the machine translation task FLORES \cite{nllb2022} to prove the effectiveness of our method on generation tasks.
Specifically, we set the following directions: \textit{en $\rightarrow$ zh}, \textit{zh $\rightarrow$ en},  \textit{en $\rightarrow$ es}, \textit{es $\rightarrow$ en},  \textit{en $\rightarrow$ el}, \textit{el $\rightarrow$ en},  \textit{en $\rightarrow$ hu}, \textit{hu $\rightarrow$ en}, \textit{en $\rightarrow$ tr}, and \textit{tr $\rightarrow$ en}. 
And we utilize COMET \cite{rei-etal-2020-comet} as the evaluation metric. 
The results in Table \ref{table:llama-res} demonstrate the effectiveness of our method across different LLMs and generative tasks, which further proves the practical application value of our method.

\begin{table*}[t]
    \centering
    \resizebox*{\linewidth}{!}{
    \begin{tabular}{l|cccccccccc|ccc}
    \bottomrule
        & \multicolumn{2}{c}{\textbf{ARC-C}} & \multicolumn{2}{c}{\textbf{MMLU}} & \multicolumn{2}{c}{\textbf{Hellaswag}} & \multicolumn{2}{c}{\textbf{Belebele}} & \multicolumn{2}{c|}{\textbf{FLORES}} & & & \\
        \hline
        \textbf{Methods}  &  \textbf{Old} &	\textbf{New} &	\textbf{Old}  &	\textbf{New} & \textbf{Old} &	\textbf{New}	& \textbf{Old}	&  \textbf{New} & \textbf{Old}	&  \textbf{New} &\textbf{\textit{Old-avg}} & \textbf{\textit{New-avg}} & \textbf{\textit{avg}}\\
        
    \hline
    Llama-3.2-3B & 44.64  & 33.39  & 	49.84 & 42.15	 &  63.56 & 43.62 & 	70.48 & 61.37 & 	84.88 & 83.96	 & 62.68 & 	52.90 & 	57.79 \\
    \hline
    MoE-LPR (3*24) & 	43.70 & 37.58 & 	49.99 & 43.98 & 	64.17 & 51.19 & 	70.74 & 64.41	 & 84.58 & 84.35 & 	62.64 & 	56.30	 & 59.47	\\
    LayerMoE (Ours)	& \textbf{44.10}	& \textbf{37.64}	& 	49.96	& \textbf{44.08}	& 	64.07	& \textbf{51.33}	& 	\textbf{71.22}	& \textbf{64.63}	& 	\textbf{84.72}	& \textbf{84.69}		& \textbf{62.81}		& \textbf{56.47}	& 	\textbf{59.64}
	\\
    \toprule
    
    \end{tabular}
    }
    \caption{
        The results of Llama-3.2-3B under the single-expansion setting (G0 \!$\rightarrow$\! G1).
        The detailed results for each language are reported in Table \ref{table:llama-res-detailed-1} and \ref{table:llama-res-detailed-2} of Appendix \ref{sec:appendix-detailed-results}.
    }
    \label{table:llama-res}
\end{table*}

\section{Related Work}
There are two categories \cite{huang2024surveylargelanguagemodels} in expanding new languages to build powerful multilingual LLMs: (1) utilizing the collected multilingual corpus to train the multilingual LLMs from scratch and (2) continually training LLMs on top of existing LLMs only with new language data.
The first category generally necessitates meticulous design in terms of the data ratio between different languages \cite{ai2024yiopenfoundationmodels, zhang2024bayling2multilinguallarge, nguyen2024seallmslargelanguage} and the implementation of curriculum learning strategies \cite{wei2023polylmopensourcepolyglot, ustun-etal-2024-aya} during the training process to ensure a balanced improvement in proficiency across various languages. 
Training from scratch typically requires excessive computational resources and training durations, which are often challenging to implement and environmentally unfriendly.

The second category is to continually train the original model with new language data to enhance the ability of new languages while avoiding the catastrophic forgetting of the ability in originally proficient languages.
MoE-LPR \cite{zhou2024moelprmultilingualextensionlarge} and MoE-CT \cite{li2024MoEctnovelapproachlarge} upcycle the dense model to the MoE model for improving the multilingual ability of LLMs.
Similarly, some methods also have introduced external parameters to improve the ability of the original LLMs and preserve the original capabilities by freezing the old parameters.
{LLaMA-Pro} \cite{wu2024llamaproprogressivellama} insert new transformer blocks to replace the duplicate layers to further enhance the original LLM.
{LoRAMoE} \cite{dou2024loraMoEalleviateworldknowledge} combines multiple low-rank adapters with a router network to achieve better performance on downstream tasks.
In this work, we only focus on the multilingual expansion and propose LayerMoE to address the limitations of existing methods, \textit{e.g.}, inefficient expansion, high resource usage, and accumulated forgetfulness of old languages.
Additionally, the concurrent work AlphaLoRA \cite{qing-etal-2024-alphalora} similar to ours allocates LoRA experts based on Heavy-Tailed Self-Regularization Theory to mitigate redundancy of LoRA experts and improve the general ability of LLMs.
Differently, we calculate the representation similarity of different languages in different layers as the indicator to allocate new experts for each layer, which is intuitive and effective for the expansion of new languages.



\section{Conclusion}
In this work, we propose a layer-wise expert allocation algorithm to assign the appropriate number of new experts for each layer when expanding new languages for existing LLMs.
We first calculate the similarity of different languages for each layer as the indicator and allocate more experts to the layers exhibiting lower similarity.
Additionally, we add a classifier in front of the router network for the layers with higher similarity to further mitigate the forgetting of old languages.
Experimental results demonstrate the effectiveness of our method, which is parameter-efficient and achieves good expansion of new languages and less forgetting of old languages.

\section*{Limitations}
Although our method achieves better improvements in terms of ``\textit{Old-avg}'' and ``\textit{New-avg}'' than MoE-LPR (6*24) with 60\% fewer experts, the improvements in different benchmarks and languages are inconsistent.
The inherent reasons for the inconsistency are not yet clear and need further investigation.
Additionally, we set the total number of new experts according to the baseline ``MoE-LPR (3*24)'' for a fair comparison rather than exploring the optimal total number.
In the future, we will continually solve these limitations for a better expert allocation solution.

\section*{Acknowledgments}
The research work described in this paper has been supported by the National Nature Science Foundation of China (No. 62476023, 61976016, 62376019, 61976015) and Henan Provincial Science and Technology Research Project (No. 252102210102), and the authors would like to thank the anonymous reviewers for their valuable comments and suggestions to improve this paper.

\bibliography{custom}

\appendix
\onecolumn
\newpage

\section{Load Balance Loss}\label{sec:appendix-load-balande-loss}
The load balance loss \cite{fedus2022switchtransformersscalingtrillion} generally mitigates the risk of routing collapse by constraining an even distribution of tokens across different experts.
Given a batch tokens $\mathbb{T}$ and $N$ experts for each layer, the load balance loss $\mathcal{L}_{balance}$ \cite{zhou2024moelprmultilingualextensionlarge} is calculated following:
\begin{equation}
    \mathcal{L}_{balance} = \sum_{i=1}^{N}f_i P_i,
\end{equation}
\begin{equation}
    f_i = \frac{N}{K|\mathbb{T}|}\sum_{t \in {\mathbb{T}}} \mathbbm{1}\{\text{Token } t \text{ selects expert }i\},
\end{equation}
\begin{equation}
    P_i = \frac{1}{|\mathbb{T}|} \sum_{t \in {\mathbb{T}}} G_i(t),
\end{equation}
where $t$ represents a token, $G_i(t)$ is the router score of the expert $i$, and $\mathbbm{1}(\cdot)$ is a indicator function when the token $t$ selecting the expert $i$ equals 1.

\section{Experimental Details}

\subsection{Detailed Introduction of Different Languages}\label{sec:appendix-languages}
The language families and writing systems of our nine selected languages are listed in Table \ref{table:languages}.
Specifically, \textit{bn}, \textit{hi}, and \textit{ne} all belong to the Indo-Aryan branch of the Indo-European family and use non-Latin scripts, making them more challenging to model effectively and less well-resourced compared to Latin-script languages.

\begin{table*}[h]
    \centering
    \resizebox*{0.8\linewidth}{!}{
    \begin{tabular}{l|ll}
    \bottomrule
        Languages	& \textbf{Language Family} & \textbf{Writing System} \\
        
    \hline

    English (\textit{en}) & Indo-European (Germanic) & Latin alphabet (26 letters) \\
    Chinese (\textit{zh}) & Sino-Tibetan (Sinitic) & Chinese characters \\
    Spanish (\textit{es}) & Indo-European (Romance) & Latin alphabet (27 letters, including ñ) \\
    Greek (\textit{el}) & Indo-European (Hellenic) & Greek alphabet (24 letters) \\
    Hungarian (\textit{hu}) & Uralic (Finno-Ugric) & Latin alphabet with diacritics (40 letters) \\
    Turkish (\textit{tr}) & Turkic & Latin alphabet (29 letters) \\
    Bengali (\textit{bn}) & Indo-European (Indo-Aryan) & Bengali script \\
    Nepali (\textit{ne}) & Indo-European (Indo-Aryan) & Devanagari script \\
    Hindi (\textit{hi}) & Indo-European (Indo-Aryan) & Devanagari script \\

    \toprule
    \end{tabular}
    }
    \caption{
        The detailed language families and writing systems.
    }
    \label{table:languages}
\end{table*}

\subsection{Evaluation Details}\label{sec:appendix-eval}

The \texttt{lm-evaluation-hardness} framework has included the multilingual version of ARC-Challenge, MMLU, and HellaSwag translated by Okapi \cite{lai-etal-2023-okapi}.
For unsupported languages in new languages, we manually integrate the Turkish version from \texttt{OpenLLM Turkish leaderboard}\footnote{\url{https://huggingface.co/spaces/malhajar/OpenLLMTurkishLeaderboard}} and the Greek version translated by ILSP\footnote{\url{https://huggingface.co/ilsp}} following the official guide\footnote{\url{https://github.com/EleutherAI/lm-evaluation-harness/blob/main/docs/new_task_guide.md}}.



\subsection{Similarity and Allocation}\label{sec:appendix-sim_and_allocation}
\begin{figure*}[h]
	\centering  
	\subfigbottomskip=2pt 
	\subfigcapskip=-5pt 
	\subfigure[G0 \!$\rightarrow$\! G1]{
		\includegraphics[width=0.32\linewidth]{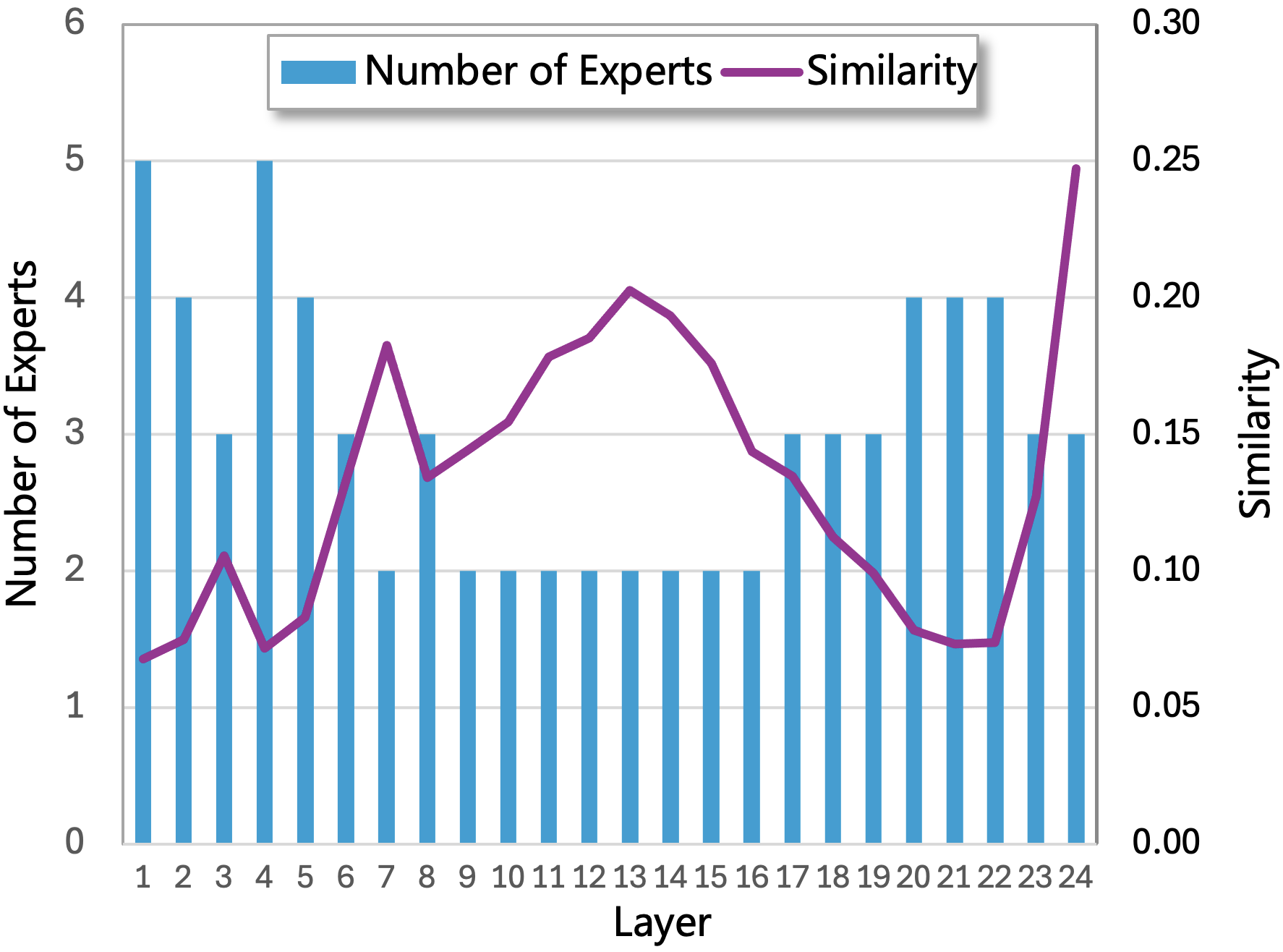}}
	\subfigure[G0 \!$\rightarrow$\! G2]{
		\includegraphics[width=0.32\linewidth]{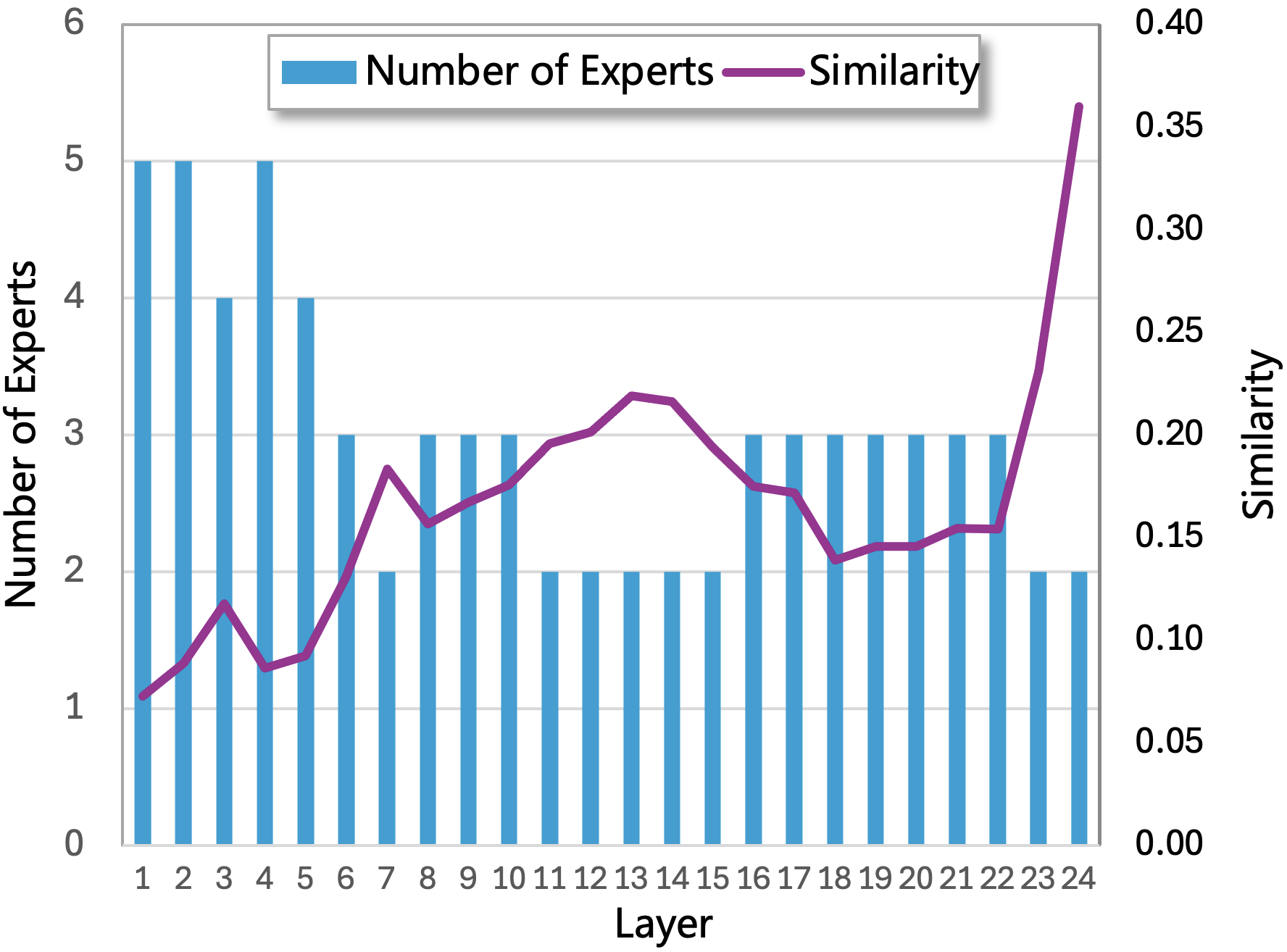}}
        \subfigure[G0 \!$\rightarrow$\! G1 \!$+$\! G2]{
		\includegraphics[width=0.32\linewidth]{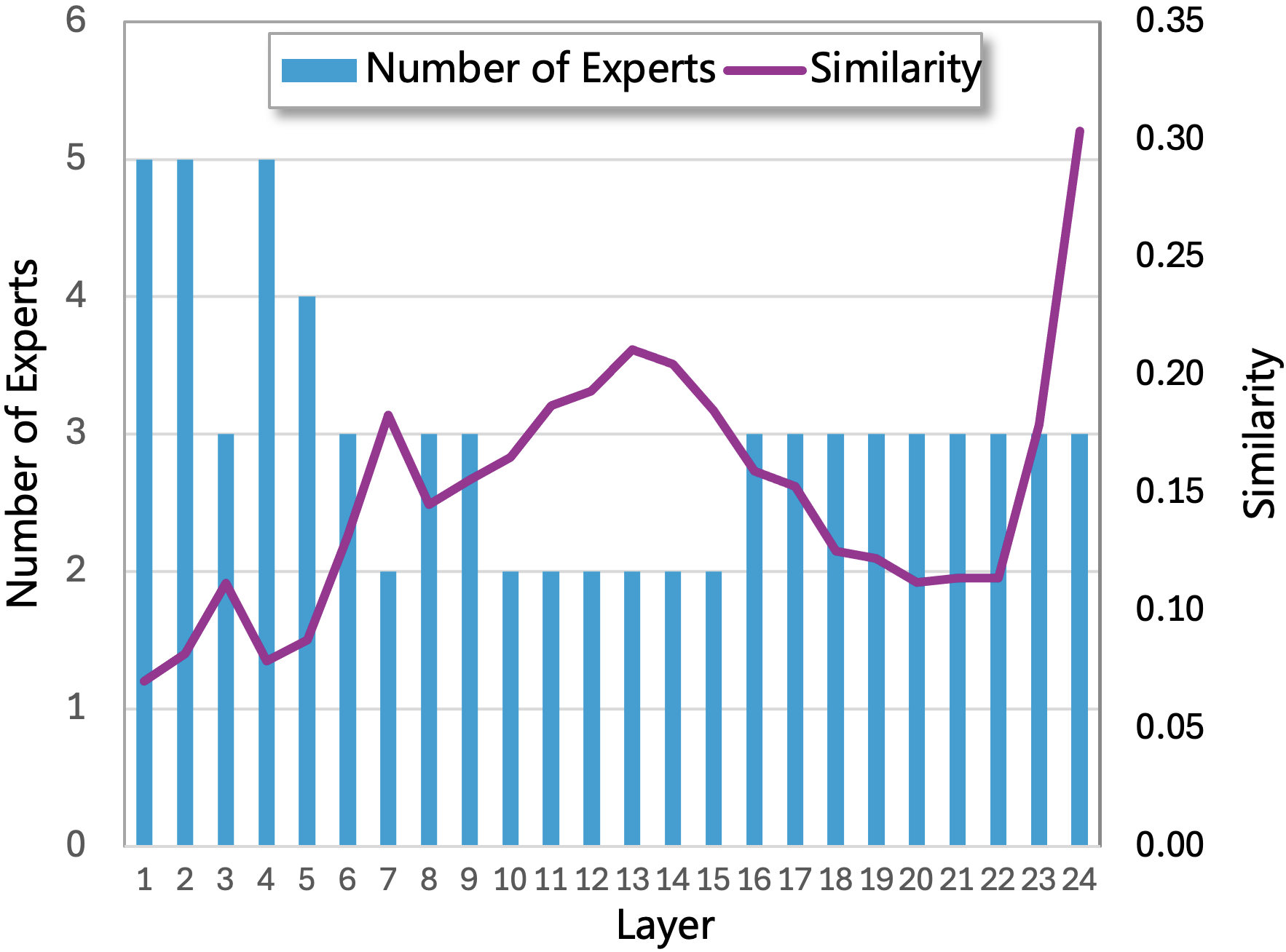}}
        \\
        \subfigure[G0 \!$\rightarrow$ \!G1 \!$\rightarrow$ \!G2]{
		\includegraphics[width=0.49\linewidth]{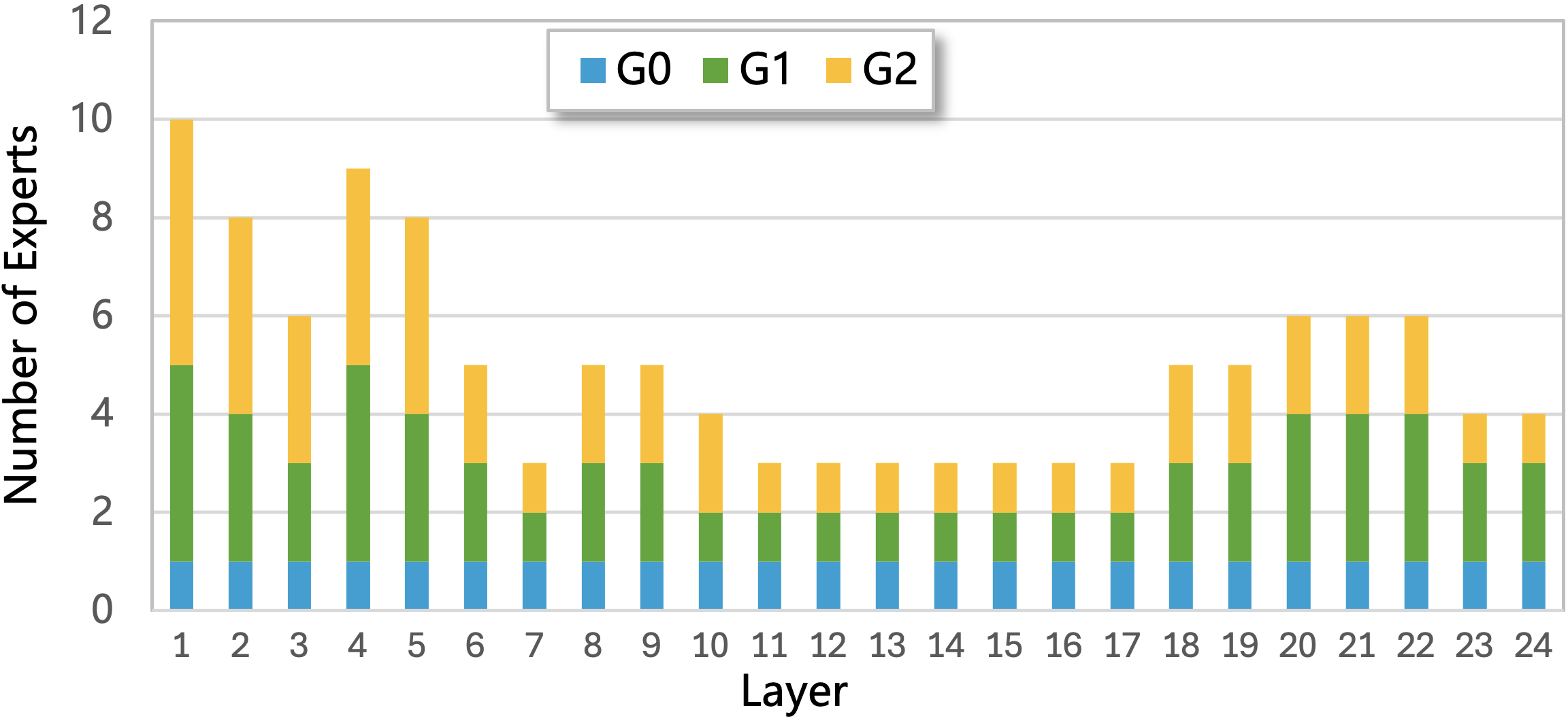}}
	\subfigure[G0 \!$\rightarrow$ \!G2 $\rightarrow$ \!G1]{
		\includegraphics[width=0.49\linewidth]{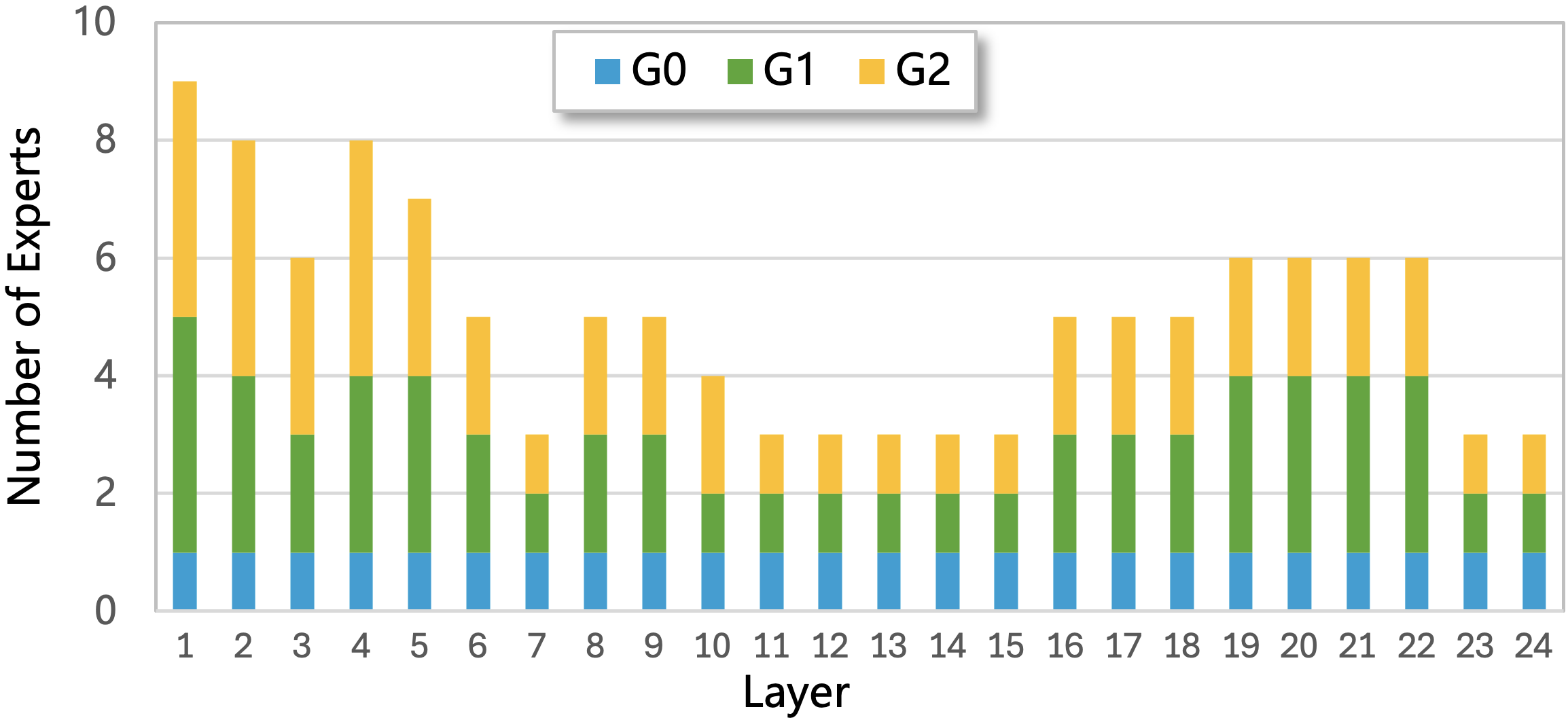}}
	\caption{The detailed similarity and expert allocation of our method for each layer of Qwen1.5-1.8B under the five settings.}
    \label{fig:sim_experts}
\end{figure*}

Figure \ref{fig:sim_experts} presents the detailed similarity and expert allocation of our method for each layer under the five settings.

\section{Detailed Results}\label{sec:appendix-detailed-results}
\paragraph{Qwen1.5-1.8B:}
The detailed results for each language under the single-expansion settings (G0 \!$\rightarrow$\! G1, G0 \!$\rightarrow$\! G2) are listed in Table \ref{table:single-res-detailed-g1} and \ref{table:single-res-detailed-g2}.
The detailed results of G0 \!$\rightarrow$\! G1 \!$+$\! G2 and the lifelong-expansion settings (G0 \!$\rightarrow$ \!G1 \!$\rightarrow$ \!G2 and G0 \!$\rightarrow$ \!G2 $\rightarrow$ \!G1) are reported in Table \ref{table:mixed-res-detailed}.

\begin{table*}[h]
    \centering
    \resizebox*{\linewidth}{!}{
    \begin{tabular}{l|ccc|ccc|ccc|ccc|c}
    \bottomrule
& \multicolumn{3}{c|}{\textbf{ARC-Challenge}} & \multicolumn{3}{c|}{\textbf{MMLU}} & \multicolumn{3}{c|}{\textbf{Hellaswag}} & \multicolumn{3}{c|}{\textbf{Belebele}} & \\
\hline
\textbf{Methods}	&	\textbf{en}	&	\textbf{es}	&	\textbf{zh}	&	\textbf{en}	&	\textbf{es}	&	\textbf{zh}	&	\textbf{en}	&	\textbf{es}	&	\textbf{zh}	&	\textbf{en}	&	\textbf{es}	&	\textbf{zh}	&	\textbf{\textit{G0-avg}}	\\
\hline
Qwen1.5-1.8B	&	37.97 	&	27.86 	&	34.10 	&	45.67 	&	38.81 	&	57.21 	&	61.62 	&	39.35 	&	48.14 	&	60.78 	&	47.44 	&	59.44 	&	46.53 	\\
\hline
MoE-LPR (6*24)	&	38.05 	&	26.84 	&	31.88 	&	45.77 	&	38.26 	&	55.50 	&	61.34 	&	39.41 	&	48.49 	&	58.44 	&	45.89 	&	54.44 	&	45.36 	\\
MOLA	&	37.88 	&	26.84 	&	32.74 	&	45.81 	&	38.30 	&	55.02 	&	61.36 	&	39.27 	&	48.26 	&	60.11 	&	47.11 	&	54.33 	&	45.59 	\\
MoE-LPR (4*24)	&	36.69 	&	27.35 	&	32.82 	&	46.09 	&	38.57 	&	54.87 	&	61.17 	&	39.37 	&	48.27 	&	58.33 	&	44.22 	&	54.44 	&	45.18 	\\
MoE-LPR (3*24)	&	37.46 	&	27.35 	&	33.50 	&	46.00 	&	38.67 	&	54.94 	&	61.20 	&	39.43 	&	48.25 	&	59.44 	&	44.44 	&	52.33 	&	45.25 	\\
LayerMoE (Ours)	&	38.57 	&	26.58 	&	33.93 	&	45.81 	&	37.97 	&	54.59 	&	60.90 	&	38.98 	&	47.84 	&	60.56 	&	46.89 	&	57.00 	&	\textbf{45.80} 	\\   
\toprule
\bottomrule
\textbf{Methods}	&	\textbf{el}	&	\textbf{hu}	&	\textbf{tr}	&	\textbf{el}	&	\textbf{hu}	&	\textbf{tr}	&	\textbf{el}	&	\textbf{hu}	&	\textbf{tr}	&	\textbf{el}	&	\textbf{hu}	&	\textbf{tr}	&	\textbf{\textit{G1-avg}}	\\
\hline
Qwen1.5-1.8B	&	22.95 	&	23.20 	&	22.63 	&	28.94 	&	32.25 	&	30.37 	&	29.15 	&	29.63 	&	29.04 	&	33.33 	&	32.00 	&	33.44 	&	28.91 	\\
\hline
MoE-LPR (6*24)	&	25.26 	&	27.14 	&	27.24 	&	32.06 	&	36.30 	&	36.49 	&	36.09 	&	38.63 	&	37.87 	&	38.33 	&	36.56 	&	42.22 	&	34.52 	\\
MOLA	&	25.94 	&	27.40 	&	27.67 	&	31.43 	&	35.80 	&	34.71 	&	35.85 	&	37.79 	&	37.46 	&	35.22 	&	39.00 	&	40.89 	&	34.10 	\\
MoE-LPR (4*24)	&	25.34 	&	27.57 	&	27.16 	&	31.58 	&	35.79 	&	34.97 	&	35.70 	&	38.26 	&	37.39 	&	34.44 	&	40.22 	&	38.56 	&	33.92 	\\
MoE-LPR (3*24)	&	24.32 	&	27.05 	&	27.92 	&	31.90 	&	36.27 	&	35.59 	&	35.55 	&	38.18 	&	37.51 	&	32.00 	&	39.56 	&	39.89 	&	33.81 	\\
LayerMoE (Ours)	&	25.34 	&	26.80 	&	27.84 	&	31.96 	&	36.76 	&	35.68 	&	35.61 	&	38.10 	&	37.31 	&	37.44 	&	41.56 	&	42.11 	&	\textbf{34.71} 	\\

    \toprule
    \end{tabular}
    }
    \caption{
        The detailed results of Qwen1.5-1.8B under the  G0 \!$\rightarrow$\! G1 setting for each language. 
    }
    \label{table:single-res-detailed-g1}
\end{table*}

\begin{table*}[ht]
    \centering
    \resizebox*{\linewidth}{!}{
    \begin{tabular}{l|ccc|ccc|ccc|ccc|c}
    \bottomrule
& \multicolumn{3}{c|}{\textbf{ARC-Challenge}} & \multicolumn{3}{c|}{\textbf{MMLU}} & \multicolumn{3}{c|}{\textbf{Hellaswag}} & \multicolumn{3}{c|}{\textbf{Belebele}} & \\
\hline
\textbf{Methods}	&	\textbf{en}	&	\textbf{es}	&	\textbf{zh}	&	\textbf{en}	&	\textbf{es}	&	\textbf{zh}	&	\textbf{en}	&	\textbf{es}	&	\textbf{zh}	&	\textbf{en}	&	\textbf{es}	&	\textbf{zh}	&	\textbf{\textit{G0-avg}}	\\
\hline
Qwen1.5-1.8B	&	37.97 	&	27.86 	&	34.10 	&	45.67 	&	38.81 	&	57.21 	&	61.62 	&	39.35 	&	48.14 	&	60.78 	&	47.44 	&	59.44 	&	46.53 	\\
\hline
MoE-LPR (6*24)	&	37.29 	&	27.26 	&	34.62 	&	46.05 	&	38.46 	&	56.60 	&	61.73 	&	39.27 	&	48.07 	&	59.67 	&	46.33 	&	53.33 	&	45.72 	\\
MOLA	&	37.37 	&	27.26 	&	32.82 	&	45.88 	&	38.71 	&	56.48 	&	61.76 	&	39.09 	&	48.34 	&	59.56 	&	46.22 	&	55.00 	&	45.71 	\\
MoE-LPR (4*24)	&	37.80 	&	27.18 	&	34.27 	&	46.00 	&	38.88 	&	56.76 	&	61.65 	&	39.03 	&	48.24 	&	59.67 	&	45.22 	&	53.67 	&	45.70 	\\
MoE-LPR (3*24)	&	37.80 	&	27.09 	&	33.42 	&	45.52 	&	38.62 	&	56.78 	&	61.59 	&	39.13 	&	48.32 	&	59.89 	&	45.33 	&	52.56 	&	45.50 	\\
LayerMoE (Ours)	&	36.86 	&	27.44 	&	33.25 	&	46.15 	&	39.05 	&	56.09 	&	61.50 	&	39.11 	&	47.77 	&	61.89 	&	46.67 	&	56.89 	&	\textbf{46.06} 	\\
\toprule
\bottomrule
\textbf{Methods}	&	\textbf{bn}	&	\textbf{hi}	&	\textbf{ne}	&	\textbf{bn}	&	\textbf{hi}	&	\textbf{ne}	& \textbf{bn}	&	\textbf{hi}	&	\textbf{ne}	& \textbf{bn}	&	\textbf{hi}	&	\textbf{ne}	&	\textbf{\textit{G2-avg}}	\\
\hline
Qwen1.5-1.8B	&	22.50 	&	23.46 	&	23.18 	&	29.59 	&	29.89 	&	29.13 	&	28.23 	&	28.48 	&	27.95 	&	29.67 	&	30.00 	&	26.89 	&	27.41 	\\
\hline
MoE-LPR (6*24)	&	24.64 	&	25.26 	&	21.64 	&	30.29 	&	30.57 	&	31.01 	&	29.90 	&	32.78 	&	30.08 	&	33.22 	&	35.22 	&	33.11 	&	29.81 	\\
MOLA	&	24.29 	&	25.09 	&	21.64 	&	31.01 	&	30.67 	&	30.77 	&	29.44 	&	32.54 	&	29.46 	&	31.00 	&	33.33 	&	32.11 	&	29.28 	\\
MoE-LPR (4*24)	&	23.52 	&	24.91 	&	21.39 	&	30.73 	&	30.54 	&	30.94 	&	30.02 	&	32.55 	&	29.45 	&	32.11 	&	34.33 	&	32.67 	&	29.43 	\\
MoE-LPR (3*24)	&	24.98 	&	24.57 	&	21.04 	&	31.59 	&	30.87 	&	29.70 	&	29.55 	&	32.68 	&	29.44 	&	32.44 	&	34.33 	&	33.22 	&	29.53 	\\
LayerMoE (Ours)	&	24.21 	&	25.26 	&	22.58 	&	31.10 	&	30.43 	&	31.01 	&	29.31 	&	32.42 	&	29.67 	&	33.44 	&	33.44 	&	35.44 	&	\textbf{29.86} 	\\
    \toprule
    \end{tabular}
    }
    \caption{
        The detailed results of Qwen1.5-1.8B under the G0 \!$\rightarrow$\! G2 setting for each language. 
    }
    \label{table:single-res-detailed-g2}
\end{table*}

\begin{table*}[ht]
    \centering
    \resizebox*{\linewidth}{!}{
    \begin{tabular}{lc|ccc|ccc|ccc|ccc|c}
    \bottomrule
& & \multicolumn{3}{c|}{\textbf{ARC-Challenge}} & \multicolumn{3}{c|}{\textbf{MMLU}} & \multicolumn{3}{c|}{\textbf{Hellaswag}} & \multicolumn{3}{c|}{\textbf{Belebele}} & \\
\hline
\textbf{Methods} & \textbf{Order}	&	\textbf{en}	&	\textbf{es}	&	\textbf{zh}	&	\textbf{en}	&	\textbf{es}	&	\textbf{zh}	&	\textbf{en}	&	\textbf{es}	&	\textbf{zh}	&	\textbf{en}	&	\textbf{es}	&	\textbf{zh}	&	\textbf{\textit{G0-avg}}	\\
\hline
Qwen1.5-1.8B & / 	&	37.97 	&	27.86 	&	34.10 	&	45.67 	&	38.81 	&	57.21 	&	61.62 	&	39.35 	&	48.14 	&	60.78 	&	47.44 	&	59.44 	&	46.53 	\\
\hline
MoE-LPR (7*24) & G0 \!$\rightarrow$\! G1 \!$+$\! G2 & 	37.88 	&	26.92 	&	33.85 	&	45.26 	&	37.61 	&	54.51 	&	61.43 	&	39.29 	&	48.14 	&	59.33 	&	44.56 	&	53.00 	&	45.15 	\\
MoE-LPR (4*24) & G0 \!$\rightarrow$\! G1 \!$+$\! G2 & 	37.80 	&	27.69 	&	32.82 	&	45.29 	&	38.12 	&	54.39 	&	61.45 	&	39.11 	&	48.20 	&	59.89 	&	44.33 	&	54.22 	&	45.28 	\\
LayerMoE (Ours) & G0 \!$\rightarrow$\! G1 \!$+$\! G2 & 	37.46 	&	27.01 	&	34.27 	&	45.63 	&	38.14 	&	54.18 	&	61.37 	&	39.08 	&	47.86 	&	60.11 	&	46.22 	&	55.44 	&	\textbf{45.56} 	\\
\hline
MoE-LPR (7*24) & G0 \!$\rightarrow$\! G1 \!$\rightarrow$\! G2 & 	37.54 	&	26.84 	&	33.16 	&	45.71 	&	38.29 	&	55.42 	&	61.43 	&	39.21 	&	48.12 	&	61.22 	&	47.33 	&	56.33 	&	45.88 	\\
LayerMoE (Ours) & G0 \!$\rightarrow$\! G1 \!$\rightarrow$\! G2 & 	37.29 	&	26.58 	&	33.93 	&	45.69 	&	38.86 	&	55.40 	&	61.23 	&	39.23 	&	48.05 	&	62.22 	&	48.00 	&	56.67 	&	\textbf{46.10} 	\\
\hline
MoE-LPR (7*24) & G0 \!$\rightarrow$\! G2 \!$\rightarrow$\! G1 & 	38.14 	&	27.18 	&	33.42 	&	46.07 	&	38.26 	&	55.28 	&	61.36 	&	39.44 	&	48.26 	&	57.44 	&	44.78 	&	53.67 	&	45.28 	\\
LayerMoE (Ours) & G0 \!$\rightarrow$\! G2 \!$\rightarrow$\! G1  & 38.31 &	27.09 &	34.70 & 46.19 & 38.79 & 55.85 & 61.19 & 39.05 & 48.05 & 61.89 & 46.44 & 57.44 & \textbf{46.25} \\

\toprule
\bottomrule
\textbf{Methods} & \textbf{Order}	&	\textbf{el}	&	\textbf{hu}	&	\textbf{tr}	&	\textbf{el}	&	\textbf{hu}	&	\textbf{tr}	&	\textbf{el}	&	\textbf{hu}	&	\textbf{tr}	&	\textbf{el}	&	\textbf{hu}	&	\textbf{tr}	&	\textbf{\textit{G1-avg}}	\\
\hline
Qwen1.5-1.8B	& / &	22.95 	&	23.20 	&	22.63 	&	28.94 	&	32.25 	&	30.37 	&	29.15 	&	29.63 	&	29.04 	&	33.33 	&	32.00 	&	33.44 	&	28.91 	\\
\hline
MoE-LPR (7*24) & G0 \!$\rightarrow$\! G1 \!$+$\! G2 & 	26.46 	&	28.00 	&	27.92 	&	30.38 	&	35.14 	&	34.64 	&	36.40 	&	39.01 	&	37.92 	&	35.33 	&	39.00 	&	39.78 	&	34.17 	\\
MoE-LPR (4*24) & G0 \!$\rightarrow$\! G1 \!$+$\! G2 & 	26.20 	&	27.65 	&	27.92 	&	31.01 	&	35.08 	&	34.37 	&	35.60 	&	38.14 	&	37.11 	&	34.11 	&	41.11 	&	40.56 	&	34.07 	\\
LayerMoE (Ours) & G0 \!$\rightarrow$\! G1 \!$+$\! G2 & 	25.34 	&	27.83 	&	27.92 	&	31.51 	&	36.43 	&	35.32 	&	35.50 	&	38.00 	&	37.27 	&	36.89 	&	42.11 	&	42.67 	&	\textbf{34.73} 	\\
\hline
MoE-LPR (7*24) & G0 \!$\rightarrow$\! G1 \!$\rightarrow$\! G2 & 	25.51 	&	26.20 	&	27.75 	&	30.51 	&	35.29 	&	34.46 	&	35.20 	&	37.92 	&	37.08 	&	37.44 	&	43.44 	&	42.00 	&	34.40 	\\
LayerMoE (Ours) & G0 \!$\rightarrow$\! G1 \!$\rightarrow$\! G2 & 	25.34 	&	25.77 	&	27.75 	&	31.94 	&	36.58 	&	35.09 	&	35.17 	&	37.92 	&	36.24 	&	37.34 	&	43.34 	&	42.18 	&	\textbf{34.56} 	\\
\hline
MoE-LPR (7*24) & G0 \!$\rightarrow$\! G2 \!$\rightarrow$\! G1 & 	25.34 	&	26.97 	&	26.47 	&	30.76 	&	34.41 	&	34.47 	&	35.11 	&	37.42 	&	37.09 	&	33.67 	&	38.44 	&	40.89 	&	33.42 	\\
LayerMoE (Ours) & G0 \!$\rightarrow$\! G2 \!$\rightarrow$\! G1  & 24.57 & 26.80 & 26.30 & 30.86	& 35.36 & 	35.26 & 34.54 &	37.07 &	36.12 & 36.44 & 39.00 & 42.78 & \textbf{33.76} \\

\toprule
\bottomrule
\textbf{Methods} & \textbf{Order}	&	\textbf{bn}	&	\textbf{hi}	&	\textbf{ne}	&	\textbf{bn}	&	\textbf{hi}	&	\textbf{ne}	& \textbf{bn}	&	\textbf{hi}	&	\textbf{ne}	& \textbf{bn}	&	\textbf{hi}	&	\textbf{ne}	&	\textbf{\textit{G2-avg}}	\\
\hline
Qwen1.5-1.8B & /	&	22.50 	&	23.46 	&	23.18 	&	29.59 	&	29.89 	&	29.13 	&	28.23 	&	28.48 	&	27.95 	&	29.67 	&	30.00 	&	26.89 	&	27.41 	\\
\hline
MoE-LPR (7*24) & G0 \!$\rightarrow$\! G1 \!$+$\! G2 & 	24.29 	&	24.91 	&	20.96 	&	30.94 	&	32.26 	&	31.31 	&	30.03 	&	33.24 	&	30.03 	&	31.56 	&	34.00 	&	36.22 	&	\textbf{29.98} 	\\
MoE-LPR (4*24) & G0 \!$\rightarrow$\! G1 \!$+$\! G2 & 	24.38 	&	25.00 	&	21.47 	&	29.64 	&	31.81 	&	30.09 	&	29.59 	&	32.72 	&	29.73 	&	33.11 	&	34.33 	&	32.78 	&	29.55 	\\
LayerMoE (Ours) & G0 \!$\rightarrow$\! G1 \!$+$\! G2 & 	23.27 	&	25.26 	&	21.47 	&	31.23 	&	32.05 	&	30.60 	&	29.67 	&	32.34 	&	29.66 	&	32.67 	&	36.00 	&	35.00 	&	29.94 	\\
\hline
MoE-LPR (7*24) & G0 \!$\rightarrow$\! G1 \!$\rightarrow$\! G2 & 	23.44 	&	25.34 	&	21.81 	&	31.66 	&	32.31 	&	31.41 	&	29.55 	&	32.60 	&	29.54 	&	34.22 	&	35.44 	&	35.78 	&	30.26 	\\
LayerMoE (Ours) & G0 \!$\rightarrow$\! G1 \!$\rightarrow$\! G2 & 	23.78 	&	25.43 	&	22.33 	&	31.42 	&	31.92 	&	31.11 	&	29.71 	&	32.49 	&	29.88 	&	34.02 	&	35.62 	&	35.82 	&	\textbf{30.29} 	\\
\hline
MoE-LPR (7*24) & G0 \!$\rightarrow$\! G2 \!$\rightarrow$\! G1 & 	25.32 	&	24.40 	&	21.81 	&	30.41 	&	31.89 	&	29.92 	&	28.79 	&	30.73 	&	29.00 	&	27.78 	&	29.67 	&	31.00 	&	28.39 	\\
LayerMoE (Ours) & G0 \!$\rightarrow$\! G2 \!$\rightarrow$\! G1 & 24.98 & 25.17 & 22.67 & 31.75 & 29.95 & 30.41 &  29.25 &	32.49 &	29.46 & 32.56 & 35.11 & 33.78 & \textbf{29.80} \\

    \toprule
    \end{tabular}
    }
    \caption{
        The detailed results of Qwen1.5-1.8B under the G0 \!$\rightarrow$\! G1 \!$+$\! G2, G0 \!$\rightarrow$ \!G1 \!$\rightarrow$ \!G2, and G0 \!$\rightarrow$ \!G2 $\rightarrow$ \!G1 settings for each language.
    }
    \label{table:mixed-res-detailed}
\end{table*}

\begin{figure}[b]
    \centering
    \includegraphics[width=0.7\linewidth]{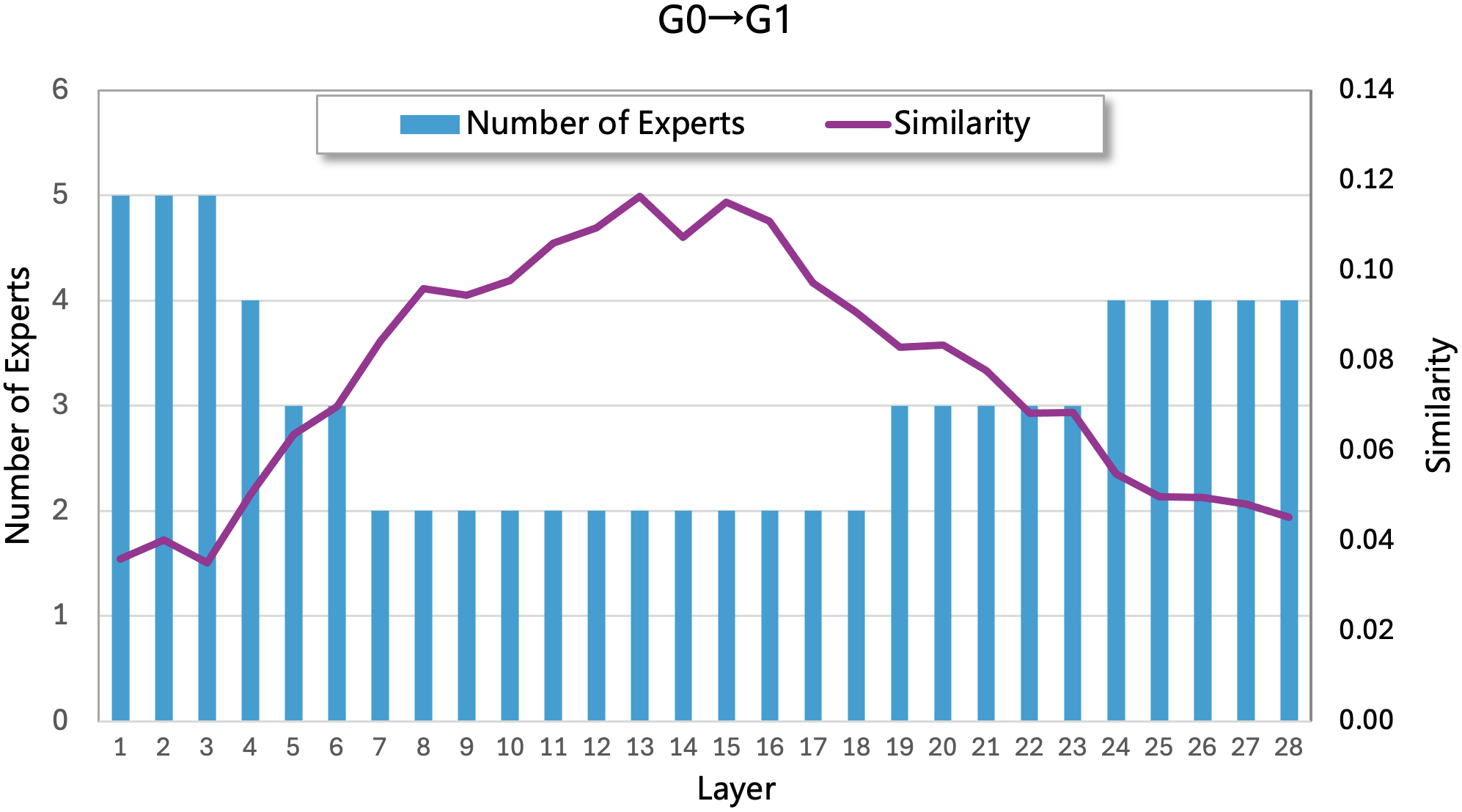}
    \caption{The detailed similarity and expert allocation of our method for each layer of Llama-3.2-3B under the  G0 \!$\rightarrow$\! G1 setting.}
    \label{fig:sim_experts_llama}
\end{figure}

\paragraph{Llama-3.2-3B:}
Figure \ref{fig:sim_experts_llama} presents the detailed similarity and expert allocation of our method for each layer of Llama-3.2-3B under the  G0 \!$\rightarrow$\! G1 setting.
The detailed results for each language under the G0 \!$\rightarrow$\! G1 setting are listed in Table \ref{table:llama-res-detailed-1} and \ref{table:llama-res-detailed-2}.

\begin{table*}[ht]
    \centering
    \resizebox*{\linewidth}{!}{
    \begin{tabular}{l|ccc|ccc|ccc|ccc|c}
    \bottomrule
& \multicolumn{3}{c|}{\textbf{ARC-Challenge}} & \multicolumn{3}{c|}{\textbf{MMLU}} & \multicolumn{3}{c|}{\textbf{Hellaswag}} & \multicolumn{3}{c|}{\textbf{Belebele}} & \\
\hline
\textbf{Methods}	&	\textbf{en}	&	\textbf{es}	&	\textbf{zh}	&	\textbf{en}	&	\textbf{es}	&	\textbf{zh}	&	\textbf{en}	&	\textbf{es}	&	\textbf{zh}	&	\textbf{en}	&	\textbf{es}	&	\textbf{zh}	&	\textbf{\textit{G0-avg}}	\\
\hline
Llama-3.2-3B	& 51.11 & 	43.59 & 	39.23  & 56.42 & 	49.09 & 	44.00 & 76.32 & 	61.50 & 	52.87 & 74.22 & 	68.56 & 	68.67 &  57.13	\\
\hline
MoE-LPR (3*24)	&	50.00  & 	42.31 & 	38.80 &  56.25 & 	49.63 & 	44.08 &  76.35 & 	62.23  & 	53.94 &  74.89 & 	69.22 & 	68.11 & 57.15 \\
LayerMoE (Ours)	&	49.91 & 	42.56 & 	39.83 &   56.05 & 	49.77 & 	44.07 &  76.28 & 	62.18 & 	53.74 &  75.33 & 	69.44 & 	68.89 & \textbf{57.34}	\\   
\toprule
\bottomrule
\textbf{Methods}	&	\textbf{el}	&	\textbf{hu}	&	\textbf{tr}	&	\textbf{el}	&	\textbf{hu}	&	\textbf{tr}	&	\textbf{el}	&	\textbf{hu}	&	\textbf{tr}	&	\textbf{el}	&	\textbf{hu}	&	\textbf{tr}	&	\textbf{\textit{G1-avg}}	\\
\hline
Llama-3.2-3B &	31.76 	&	34.85 	&	33.56 	&	41.08 	&	43.01 	&	42.37 	&	43.37 	&	44.65 	&	42.85 	&	65.22 	&	59.67 	&	59.22 	& 45.13	\\
\hline
MoE-LPR (3*24)	&	36.64 	&	39.21 	&	36.89 	&	43.96 	&	44.51 	&	43.46 	&	52.49 	&	51.33 	&	49.74 	&	68.44 	&	63.00 	&	61.78 	& 	49.29 \\
LayerMoE (Ours)	&	36.64 	&	39.73 	&	36.55 	&	43.88 	&	44.86 	&	43.51 	&	52.72 	&	51.66 	&	49.60 	&	67.89 	&	64.22 	&	61.78 	&	\textbf{49.42} \\

    \toprule
    \end{tabular}
    }
    \caption{
        The results of Llama-3.2-3B on ARC-Challenge, MMLU, Hellaswag, and Belebele under the G0 \!$\rightarrow$\! G1 setting for each language. 
    }
    \label{table:llama-res-detailed-1}
\end{table*}

\begin{table*}[ht]
    \centering
    \resizebox*{\linewidth}{!}{
    \begin{tabular}{l|cccc|c|cccccc|c}
    \bottomrule
\textbf{Methods}	&	\textbf{\textit{en $\rightarrow$ zh}}	&	\textbf{\textit{zh $\rightarrow$ en}}	&	\textbf{\textit{en $\rightarrow$ es}}	&	\textbf{\textit{es $\rightarrow$ en}}	& \textbf{old-avg}	 & \textbf{\textit{en $\rightarrow$ el}}	&	\textbf{\textit{el $\rightarrow$ en}}	&	\textbf{\textit{en $\rightarrow$ hu}}	&	\textbf{\textit{hu $\rightarrow$ en}}	& \textbf{\textit{en $\rightarrow$ tr}}	&	\textbf{\textit{tr $\rightarrow$ en}}	& \textbf{new-avg} \\
\hline
Llama-3.2-3B	&  84.36	 & 83.95 & 	84.79	 & 86.4	 & 84.88 & 	82.47	 & 85.86	 & 83.81 & 	86.15	 & 78.97	 & 86.53 & 	83.96 \\
\hline
MoE-LPR (3*24)	&	84.19 & 	83.19 & 	84.67 & 	86.27 & 	84.58	 & 84.18	 & 85.89	 & 83.24	 & 85.6	 & 81.95	 & 85.26 & 	84.35  \\
LayerMoE (Ours)	&	84.32 & 	83.67	 & 84.63 & 	86.26	 & \textbf{84.72}	 & 85.14	 & 86.18	 & 83.7	 & 85.65 & 	82.02 & 	85.44 & 	\textbf{84.69}	\\   
\toprule
    \end{tabular}
    }
    \caption{
        The detailed results of Llama-3.2-3B on the FLORES benchmark under the G0 \!$\rightarrow$\! G1 setting for each language. 
    }
    \label{table:llama-res-detailed-2}
\end{table*}

\end{document}